\documentclass[11pt]{article}

\usepackage[numbers,sort]{natbib}
\usepackage[utf8]{inputenc} 
\usepackage[T1]{fontenc}    
\usepackage[english]{babel}
\usepackage[in]{fullpage}
\usepackage{hyperref}       
\usepackage{url}            
\usepackage{booktabs}       
\usepackage{amsfonts}       
\usepackage{nicefrac}       
\usepackage{microtype}      
\usepackage{xcolor}         
\usepackage{parskip}
\usepackage{enumitem}
\usepackage{placeins}
\usepackage{xargs}
\usepackage[colorinlistoftodos,prependcaption,textsize=small]{todonotes}
\usepackage{makecell}
\usepackage{etoc}

\newcommandx{\LS}[2][1=]{\todo[linecolor=blue,backgroundcolor=blue!25,bordercolor=blue,#1]{LS: #2}}
\newcommandx{\john}[2][1=]{\todo[linecolor=green,backgroundcolor=green!25,bordercolor=green,#1]{John: #2}}

\title{Retiring Adult: New Datasets for Fair Machine Learning}

\author{
    Frances Ding\thanks{Authors ordered alphabetically}\\
    UC Berkeley
    \and
    Moritz Hardt\footnotemark[1]\\
    UC Berkeley
    \and
    John Miller\footnotemark[1]\\
    UC Berkeley
    \and
    Ludwig Schmidt\footnotemark[1]\\
    Toyota Research Institute
}

\date{}

\begin{document}

\maketitle

\begin{abstract}
Although the fairness community has recognized the importance of data,
researchers in the area primarily rely on UCI Adult when it comes to tabular
data. Derived from a 1994 US Census survey, this dataset has appeared in
hundreds of research papers where it served as the basis for the development
and comparison of many algorithmic fairness interventions. We reconstruct a
superset of the UCI Adult data from available US Census sources and reveal
idiosyncrasies of the UCI Adult dataset that limit its external validity. Our
primary contribution is a suite of new datasets derived from US Census surveys
that extend the existing data ecosystem for research on fair machine learning.
We create prediction tasks relating to income, employment, health,
transportation, and housing. The data span multiple years and all states of the
United States, allowing researchers to study temporal shift and geographic
variation. We highlight a broad initial sweep of new empirical insights
relating to trade-offs between fairness criteria, performance of algorithmic
interventions, and the role of distribution shift based on our new datasets.
Our findings inform ongoing debates, challenge some existing narratives, and
point to future research directions.
\end{abstract}

\etocdepthtag.toc{mtsection}
\section{Introduction}
\label{sec:intro}

Datasets are central to the machine learning ecosystem. Besides providing
training and testing data for model builders, datasets formulate problems,
organize communities, and interface between academia and industry. Influential
works relating to the ethics and fairness of machine learning recognize the
centrality of datasets, pointing to significant harms associated with data, as
well as better data practices \citep{buolamwini2018gender, gebru2018datasheets,
jo2020lessons, onuoha2016point, paullada2020data}. 
While the discourse about data has prioritized
cognitive domains such as vision, speech, or language, numerous consequential
applications of predictive modeling and risk assessment involve bureaucratic, organizational, and administrative records best represented as tabular data \citep{pasquale2015black, eubanks2018automating, benjamin2019race}.

When it comes to tabular data, surprisingly, most research papers on
algorithmic fairness continue to involve a fairly limited collection of
datasets, chief among them the \emph{UCI Adult} dataset \citep{kohavi1996uci}. 
Derived from the 1994 Current
Population Survey conducted by the US Census Bureau, this dataset has made an
appearance in more than three hundred research papers related to fairness where
it served as the basis for the development and comparison of many algorithmic
fairness interventions.

Our work begins with a critical examination of the UCI Adult dataset---its
origin, impact, and limitations. To guide this investigation we identify the
previously undocumented exact source of the UCI Adult dataset, allowing us to
reconstruct a superset of the data from available US Census records. This
reconstruction reveals a significant idiosyncrasy of the UCI Adult prediction
task that limits its external validity. 

While some issues with UCI Adult are readily apparent, such as its age, limited
documentation, and outdated feature encodings, a significant problem may be
less obvious at first glance. Specifically, UCI Adult has a binary target label
indicating whether the income of a person is greater or less than fifty
thousand US dollars. This income threshold of \$50k US dollars corresponds to
the 76th quantile of individual income in the United States in 1994, the 88th
quantile in the Black population, and the 89th quantile among women. We show
how empirical findings relating to algorithmic fairness are sensitive to the
choice of the income threshold, and how UCI Adult exposes a rather extreme
threshold. Specifically, the magnitude of violations in different fairness
criteria, trade-offs between them, and the effectiveness of algorithmic
interventions all vary significantly with the income threshold. In many cases,
the \$50k threshold understates and misrepresents the broader picture.

Turning to our primary contribution, we provide a suite of new datasets derived
from US Census data that extend the existing data ecosystem for research on
fair machine learning. These datasets are derived from two different data
products provided by the US Census Bureau. One is the Public Use Microdata
Sample of the American Community Survey, involving millions of US households
each year. The other is the Annual Social and Economic Supplement of the
Current Population Survey. Both released annually, they represent major
surveying efforts of the Census Bureau that are the basis of important policy
decisions, as well as vital resources for social scientists.

We create prediction tasks in different domains, including income, employment,
health, transportation, and housing. The datasets span multiple years and all
states of the United States, in particular, allowing researchers to study
temporal shift and geographic variation. 
Alongside these prediction tasks, we release a Python package called {\tt
folktables} which interfaces with Census data sources and allows users to both
access our new predictions tasks and create new tasks from Census data through
a simple API\footnote{The datasets and Python package are available for download at
\url{https://github.com/zykls/folktables}.}.

We contribute a broad initial sweep of new empirical insights into algorithmic
fairness based on our new datasets. Our findings inform ongoing debates and in
some cases challenge existing narratives about statistical fairness criteria
and algorithmic fairness interventions. We highlight three robust observations:
\begin{enumerate}
\item Variation within the population plays a major role in empirical observations and how they should be interpreted:
\begin{enumerate}[label=(\alph*)]
    \item Fairness criteria and the effect size of different interventions varies greatly by state. This shows that statistical claims about algorithmic fairness must be qualified carefully by context, even though they often are not.
    \item Training on one state and testing on another generally leads to unpredictable results. Accuracy and fairness criteria could change in either direction. This shows that algorithmic tools developed in one context may not transfer gracefully to another.
    \item Somewhat surprisingly, fairness criteria appear to be more stable over time than predictive accuracy. This is true both before and after intervention.
\end{enumerate}
\item Algorithmic fairness interventions must specify a locus of intervention. For example, a model could be trained on the entire US population, or on a state-by-state basis. The results differ significantly. Recognition of the need for such a choice is still lacking, as is scholarship guiding the practitioner on how to navigate this choice and its associated trade-offs.
\item Increased dataset size does not necessarily help in reducing observed disparities. Neither does social progress as measured in years passed. This is in contrast to intuition from cognitive machine learning tasks where more representative data can improve metrics such as error rate disparities between different groups.
\end{enumerate}

Our observations apply to years of active research into algorithmic fairness,
and our work provides new datasets necessary to re-evaluate and extend the
empirical foundations of the field.

\section{Archaeology of UCI Adult: Origin, Impact, Limitations}
\label{sec:archaeology}

\begin{quote}\footnotesize\it
Archaeology organises the past to understand the present. It lifts the
dust-cover off a world that we take for granted. It makes us reconsider what we
experience as inevitable.

\hfill--- Ian Hacking
\end{quote}

Although taken for granted today, the use of benchmark datasets in machine
learning emerged only in late 1980s \citep{hardtrecht}. Created in 1987, the UCI Machine Learning
Repository contributed to this development by providing researchers with
numerous datasets each with a fixed training and testing split \citep{langley2011changing}. As of writing,
the UCI Adult dataset is the second most popular dataset among more than five
hundred datasets in the UCI repository. An identical dataset is called ``Census
Income Data Set'' and a closely related larger dataset goes by ``Census-Income
(KDD) Data Set''.

At the outset, UCI Adult contains 48,842 rows each apparently describing one
individual with 14 attributes. The dataset information reveals that it
was extracted from the ``1994 Census database'' according to certain filtering
criteria. Since the US Census Bureau provides several data products, as we will
review shortly, this piece of information does not identify the source of the
dataset.

The fourteen features of UCI Adult include what the fairness community calls
\emph{sensitive} or \emph{protected} attributes, such as, age, sex, and
race. The earliest paper on algorithmic fairness that used UCI Adult to our
knowledge is a work by Calders et al.~\cite{calders2009building} from 2009. The
availability of sensitive attributes contributed to the choice of the dataset
for the purposes of this work. An earlier paper in this context by Pedreschi et al.~\cite{pedreschi2008discrimination} used the UCI German credit dataset, which is
smaller and ended up being less widely used in the community. Another highly
cited paper on algorithmic fairness that popularized UCI Adult is the work of 
Zemel et al.~\cite{zemel2013learning} on \emph{learning fair representations} (LFR). 
Published in 2013, the work introduced the idea of changing the data
representation to achieve a particular fairness criterion, in this case,
demographic parity, while representing the original data as well as possible.
This idea remains popular in the community and the LFR method has become a
standard baseline.

Representation learning is not the only topic for which UCI Adult became the
standard test case. The dataset has become broadly used throughout the area for
purposes including the development of new fairness criteria, algorithmic
interventions and fairness promoting methods, as well as causal modeling. Major
software packages, such as AI Fairness 360 \citep{bellamy2019ai} and Fairlearn 
\citep{bird2020fairlearn}, expose UCI Adult as
one of a few standard examples. Indeed, based on bibliographic information
available on Google Scholar there appear to be more than 300 papers related to
algorithmic fairness that used the UCI Adult dataset at the time of writing.

\subsection{Reconstruction of UCI Adult}
\label{sec:reconstruction}
Creating a dataset involves a multitude of design choices that substantially affect the validity of experiments conducted with the dataset.
To fully understand the context of UCI Adult and explore variations of its design choices, we reconstructed a closely matching superset from the original Census sources.
We now describe our reconstruction in detail and then investigate one specific design choice, the income binarization threshold, in Section \ref{sec:income_threshold}.

The first step in our reconstruction of UCI Adult was identifying the original data source.
As mentioned above, the ``1994 census database`` description in the UCI Adult documentation does not uniquely identify the data product provided by the US Census Bureau.
Based on the documentation of the closely related ``Census-Income (KDD) Data
Set,''\footnote{Ron Kohavi is a co-creator of both datasets.} we decided to
start with the Current Population Survey (CPS) data, specifically the Annual Social and Economic Supplement (ASEC) from 1994.
We utilized the IPUMS interface to the CPS data \citep{ipums} and hence refer to our reconstruction as IPUMS Adult.

The next step in the reconstruction was matching the 15 features in UCI Adult to the CPS data.
This was a non-trivial task: the UCI Adult documentation does not mention any specific CPS variable names and IPUMS CPS contains more than 400 candidate variables for the 1994 ASEC.
To address this challenge, we designed the following matching procedure that we repeated for each feature in UCI Adult: First, identify a set of candidate variables in CPS via the IPUMS keyword search. For each candidate variable, use the CPS documentation to manually derive a mapping from the CPS encoding to the UCI Adult encoding.
Finally, match each row in UCI Adult to its nearest neighbor in the partial reconstruction assembled from previous exact variable matches.

We only included a candidate variable if the nearest neighbor match was \emph{exact}, i.e., we could find an exact match in the IPUMS CPS data for each row in UCI Adult that matched \emph{both} the candidate variable and all earlier variables also identified via exact matches. There were only two exceptions to this rule. We discuss them in Appendix~\ref{app:adult_reconstruction}.
After completing the variable matching, our reconstruction has 49,531 rows when we use the same inclusion criteria as UCI Adult to the extent possible, which is slightly more than the 48,842 rows in UCI Adult.
The discrepancy likely stems from the fact that UCI Adult used the variable ``fnlwgt'' in its inclusion criteria and we did not due to the lack of an exact match for this variable.
This made our inclusion criteria slightly more permissive than those of UCI Adult.
The fact that we found exact matches for 13 of the 15 UCI Adult variables and a very close match for ``native-country'' is evidence that our reconstruction of UCI Adult is accurate.

\begin{figure*}
    \centering
    \includegraphics[width=\linewidth]{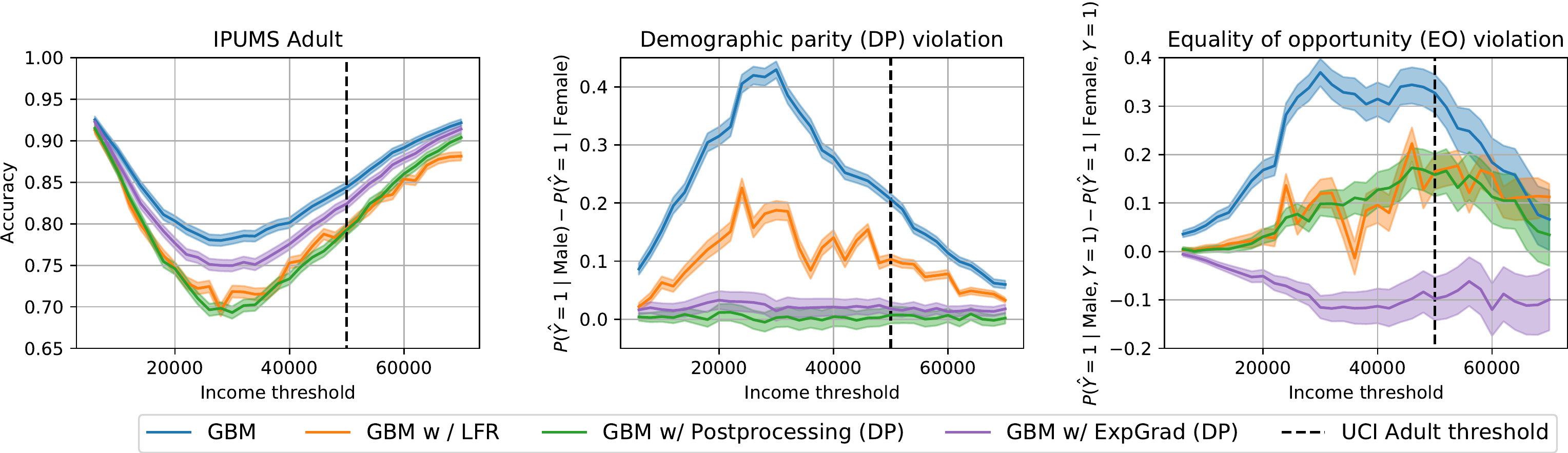}
    \caption{
        Fairness interventions with varying income threshold on IPUMS Adult.
        We compare three methods for achieving demographic parity: a
        pre-processing method (LFR), an in-training method based
        on~\citet{agarwal2018reductions} (ExpGrad), and a post-processing
        adjustment method~\cite{hardt2016equality}.  We apply each method using
        a gradient boosted decision tree (GBM) as the base classifier.
        Confidence intervals are 95\% Clopper-Pearson intervals for accuracy and
        95\% Newcombe intervals for DP.
    }
    \label{fig:adult_vary_threshold}
\end{figure*}

\subsection{Varying income threshold}
\label{sec:income_threshold}
The goal in the UCI Adult dataset is to predict whether an individual earns
greater than 50,000 US dollars a year. The choice of the 50,000 dollar threshold
is idiosyncratic and potentially limits the external validity of UCI Adult as a
benchmark for algorithmic fairness. 
In 1994, the median US income was 26,000 dollars,
and 50,000 dollars corresponds to the 76th quantile of the income distribution,
and the 88th and 89th quantiles of the income distribution for the Black and
female populations, respectively. Consequently, \emph{almost all of the Black
and female instances in the dataset fall below the threshold} and  models
trained on UCI adult tend to have substantially higher accuracies on these
subpopulations. For instance, a standard logistic regression model trained on
UCI Adult dataset achieves 85\% accuracy overall, 91.4\% accuracy on the Black
instances, and 92.7\% on Female instances. This is a rather untypical situation
since often machine learning models perform more poorly on historically
disadvantaged groups.

To understand the sensitivity of the empirical findings on UCI Adult to the
choice of threshold, we leverage our IPUMS Adult reconstruction, which includes the
continuous, unthresholded income variable, and construct a new collection of
datasets where the income threshold varies from 6,000 to 70,000.
For each threshold, we first train a standard
gradient boosted decision tree and evaluate both its accuracy and its
violation of two common fairness criteria: \emph{demographic parity} (equality of positive rates) and
\emph{equal opportunity} (equality of true positive rates). See the text
by~\citet{barocas-hardt-narayanan} for background.  The results are presented in
Figure~\ref{fig:adult_vary_threshold}, where we see both accuracy and the
magnitude of violations of these criteria vary substantially with the threshold
choice. 

We then evaluate how the choice of threshold affects three common classes of
fairness interventions: the preprocessing method LFR~\citep{zemel2013learning} mentioned earlier, an \emph{in-processing} or \emph{in-training} method based on the reductions approach
in~\citet{agarwal2018reductions}, and the post-processing method
from~\citet{hardt2016equality}.  In Figure~\ref{fig:adult_vary_threshold}, we
plot model accuracy after applying each intervention to achieve demographic
parity as well as violations of both demographic parity and equality of
opportunity as the income threshold varies. In
Appendix~\ref{app:adult_reconstruction}, we conduct the same experiment for
methods to achieve equality of opportunity.  There are three salient findings.
First, the effectiveness of each intervention depends on the threshold. For
values of the threshold near 25,000, the accuracy drop needed to achieve
demographic parity or equal opportunity is significantly larger than closer to
50,000. Second, the trade-offs between different criteria vary
substantially with the threshold. Indeed, for the in-processing method enforcing
demographic parity, as the threshold varies, the equality of opportunity
violation is monotonically increasing. Third, for high values of the threshold,
the small number of positive instances substantially enlarges the
confidence intervals for equality of opportunity, which makes it difficult to
meaningfully compare the performance of methods for satisfying this constraint.

\section{New datasets for algorithmic fairness}
\label{sec:new_data}

At least one aspect of UCI Adult is remarkably positive. The US Census Bureau invests heavily in high quality data collection, surveying methodology, and documentation based on decades of experience. Moreover, responses to some US Census Bureau surveys are legally mandated and hence enjoy high response rates resulting in a representative sample. In contrast, some notable datasets in machine learning are collected in an ad-hoc manner, plagued by skews in representation \citep{torralba2011unbiased, bolukbasi2016man, caliskan2017semantics, yang2020towards}, often lacking copyright \citep{levendowski2018copyright} or consent from subjects \citep{prabhu2020large}, and involving unskilled or poorly compensated labor in the form of crowd workers \citep{gray2019ghost}.

In this work, we tap into the vast data ecosystem of the US Census Bureau to create new machine learning tasks that we hope help to establish stronger empirical evaluation practices within the algorithmic fairness community.

As previously discussed, UCI Adult was derived from the Annual Social and Economic Supplement (ASEC) of the Current Population Survey (CPS). The CPS is a monthly survey of approximately 60,000 US households. It's used to produce the official monthly estimates of employment and unemployment for the United States. The ASEC contains additional information collected annually.

Another US Census data product most relevant to us are the American Community Survey (ACS) Public Use Microdata Sample (PUMS). ACS PUMS differs in some significant ways from CPS ASEC. The ACS is sent to approximately 3.5 million US households each year gathering information relating to ancestry, citizenship, education, employment, language proficiency, income, disability, and housing characteristics. Participation in the ACS is mandatory under federal law. Responses are confidential and governed by strict privacy rules. The Public Use Microdata Sample contains responses to every question from a subset of respondents. The geographic information associated with any given record is limited to a level that aims to prevent re-identification of survey participants. A number of other disclosure control heuristics are implemented. Extensive documentation is available on the websites of the US Census Bureau.

\begin{table}
  \caption{New prediction task details instantiated on 2018 US-wide ACS PUMS data}
  \label{table:tasks}
  \centering
  \begin{tabular}{lcllll}
    \toprule
      Task & Features & Datapoints & \makecell{Constant \\ predictor acc} & LogReg acc & GBM acc \\
    \midrule
        ACSIncome & 10 & 1,664,500 & 63.1\% & 77.1\% & 79.7\% \\
        ACSPublicCoverage & 19 & 1,138,289 & 70.2\% & 75.6\% & 78.5 \%\\
        ACSMobility & 21 & 620,937 & 73.6\% & 73.7\% &  75.7\% \\
        ACSEmployment & 17 & 3,236,107 & 56.7\% & 74.3\% & 78.5\% \\
        ACSTravelTime & 16 & 1,466,648 & 56.3\% & 57.4\% &  65.0\% \\
    \bottomrule
  \end{tabular}
\end{table}

\subsection{Available prediction tasks}
We use ACS PUMS as the basis for the following new prediction tasks:

\textbf{ACSIncome:} predict whether an individual's income is above \$50,000, after filtering the ACS PUMS data sample to only include individuals above the age of 16, who reported usual working hours of at least 1 hour per week in the past year, and an income of at least \$100. The threshold of \$50,000 was chosen so that this dataset can serve as a replacement to UCI Adult, but we also offer datasets with other income cutoffs described in Appendix~\ref{appendix:new-tasks}.

\textbf{ACSPublicCoverage:} predict whether an individual is covered by public health insurance, after filtering the ACS PUMS data sample to only include individuals under the age of 65, and those with an income of less than \$30,000. This filtering focuses the prediction problem on low-income individuals who are not eligible for Medicare.

\textbf{ACSMobility:} predict whether an individual had the same residential address one year ago, after filtering the ACS PUMS data sample to only include individuals between the ages of 18 and 35. This filtering increases the difficulty of the prediction task, as the base rate of staying at the same address is above 90\% for the general population. 

\textbf{ACSEmployment:} predict whether an individual is employed, after filtering the ACS PUMS data sample to only include individuals between the ages of 16 and 90. 

\textbf{ACSTravelTime:} predict whether an individual has a commute to work that is longer than 20 minutes, after filtering the ACS PUMS data sample to only include individuals who are employed and above the age of 16. The threshold of 20 minutes was chosen as it is the US-wide median travel time to work  in the 2018 ACS PUMS data release.

All our tasks contain features for age, race, and sex, which correspond to
\emph{protected categories} in different domains under US anti-discrimination
laws~\cite{barocas2016big}. Further, each prediction task can be instantiated on
different ACS PUMS data samples, allowing for comparison across geographic and
temporal variation. We provide datasets for each task corresponding to 1) all
fifty US states and Puerto Rico, and 2) five different years of data collection:
2014--2018 inclusive, resulting in a total of 255 distinct datasets per task to
assess distribution shift. We also provide US-wide datasets for each task,
constructed from concatenating  each state's data. Table \ref{table:tasks}
displays more details about each prediction task as instantiated on the 2018
US-wide ACS PUMS data sample. Our new tasks constitute a diverse collection of
prediction problems ranging from those where machine learning achieves
significantly higher accuracy than a baseline constant predictor to other
potentially low-signal problems (ACSMobility) where accuracy improvement appears
to be more challenging.  We also provide the exact features included in each
prediction task, and other details, in Appendix \ref{appendix:new-tasks}.
A datasheet~\cite{gebru2018datasheets} for our datasets is provided in
Appendix~\ref{sec:Datasheet}. 

These prediction tasks are by no means exhausitive of the potential tasks one
can construct using the ACS PUMS data. The {\tt folktables} package we introduce
provides a simple API that allows users to construct new tasks using the ACS
PUMS data, and we encourage the community to explore additional prediction tasks
beyond those introduced in this paper.

\subsection{Scope and limitations}
\label{sec:limitations}

One distinction is important. Census data is often used by social scientists to study the extent of inequality in income, employment, education, housing or other aspects of life. Such important substantive investigations should necessarily inform debates about discrimination in classification scenarios within these domains. However, our contribution is not in this direction. We instead use census data for the empirical study of algorithmic fairness. This generally may include performance claims about specific methods, the comparison of different methods for achieving a given fairness metric, the relationships of different fairness criteria in concrete settings, causal modeling of different scenarios, and the ability of different methods to transfer successfully from one context to another. We hope that our work leads to more comprehensive empirical evaluations in research papers on the topic, at the very least reducing the overreliance on UCI Adult and providing a complement to the flourishing theoretical work on the topic. The distinction we draw between benchmark data and substantive domain-specific investigations resonates with recent work that points out issues with using data about risk assessments tools from the criminal justice domain as machine learning benchmarks~\cite{bao2021s}.

A notable if obvious limitation of our work is that it is entirely US-centric. A richer dataset ecosystem covering international contexts within the algorithmic fairness community is still lacking. Although empirical work in the Global South is central in other disciplines, there continues to be much need for the North American fairness community to engage with it more strongly \citep{abebe2021narratives}.

\section{A tour of empirical observations}
\label{sec:experiments}

\begin{figure*}
    \centering
    \includegraphics[width=\linewidth]{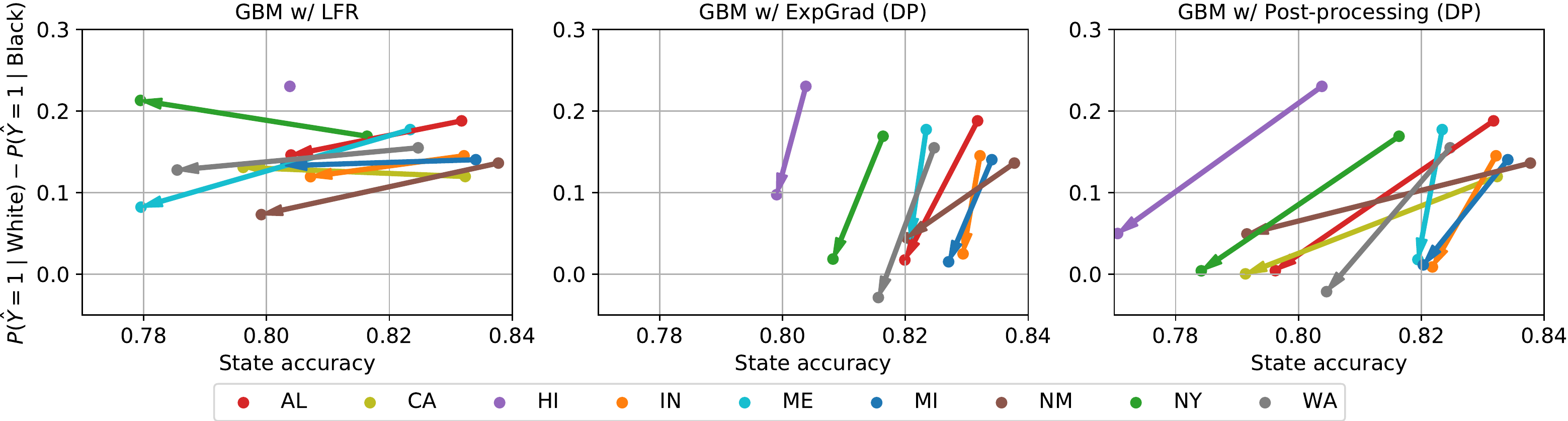}
    \caption{
        The effect size of fairness interventions varies by state. Each panel
        shows the change in accuracy and demographic parity on the ACSIncome
        task after applying a fairness intervention to an unconstrained
        gradient boosted decision tree (GBM). Each arrow corresponds to a different
        state distribution. The arrow base represents the (accuracy, DP) point
        corresponding to the unconstrained GBM, and the head represents the
        (accuracy, DP) point obtained after applying the intervention. The
        arrow for HI in the LFR plot is entirely covered by the start and end
        points.
    }
    \label{fig:PUMSAdult_fairness_flow}
\end{figure*}

In this section, we highlight an initial sweep of empirical observations enabled
by our new ACS PUMS derived prediction tasks. Our experiments focus on three
fundamental issues in fair machine learning: (i) variation within the population
of interest, e.g., how does the effectiveness of interventions vary between
different states or over time?, (ii) the locus of intervention, e.g. should
interventions be performed at the state or national level?, and (iii) whether
increased dataset size or the passage of time mitigates observed disparities?

Our experiments are not exhaustive and are intended to highlight the perspective
a broader empirical evaluation with our new datasets can contribute to
addressing questions within algorithmic fairness. The goal of the experiments is
not to provide a complete overview of all the questions that one can answer
using our datasets.  Rather, we hope to inspire other researchers to creatively
use our datasets to further probe these question as well as propose new ones
leveraging the ACS PUMS data.

\subsection{Variation within the population}
The ACS PUMS prediction tasks present two natural axes of variation: geographic
variation between states and temporal variation between years the ACS is
conducted. This variation allows us to both measure the performance of different
fairness interventions on a broad collection of different distributions, as well
as study the performance of these interventions under geographical and
temporal \emph{distribution shift} when the test dataset differs from the one on
which the model was trained. 

\begin{figure*}
    \centering
    \includegraphics[width=\linewidth]{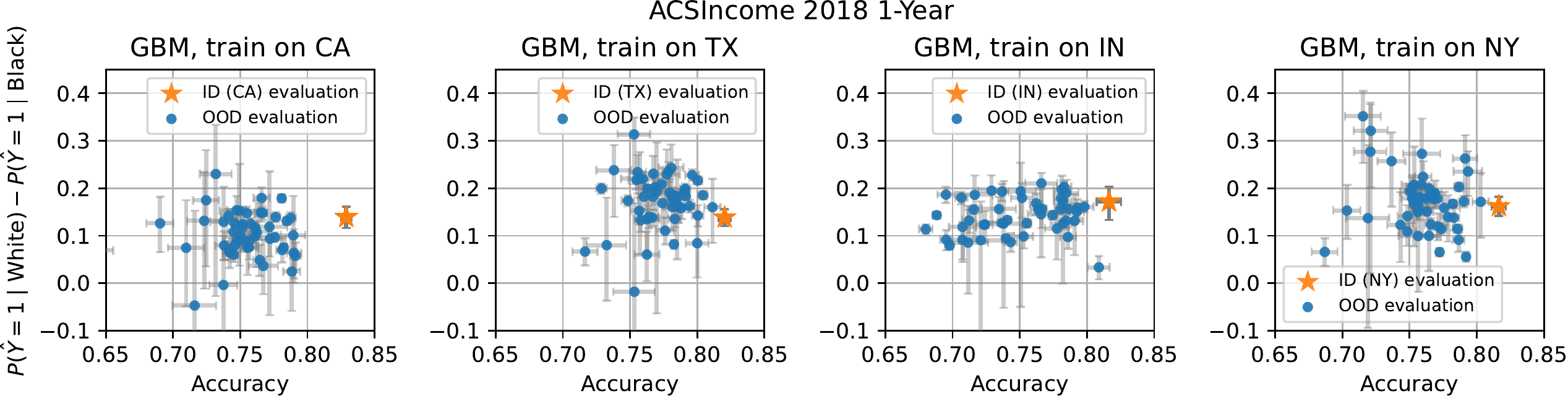}
    \includegraphics[width=\linewidth]{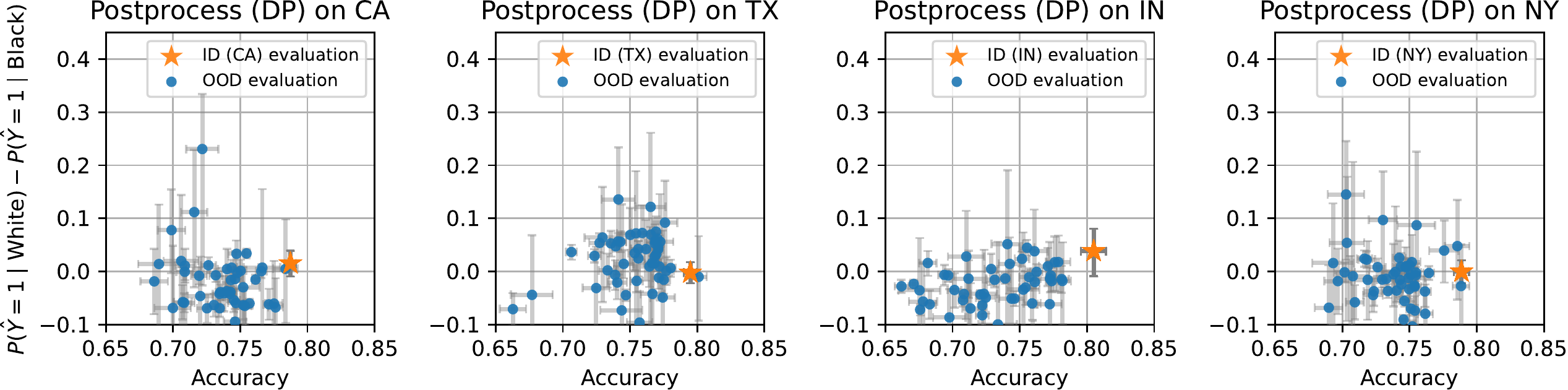}
    \caption{
        Transfer from one state to another gives unpredictable results in terms
        of predictive accuracy and fairness criteria. \textbf{Top:} Each panel
        shows an unconstrained GBM trained on a particular state on the
        ACSIncome task and evaluated both in-distribution (ID) on the same
        state and out-of-distribution (OOD) on the 49 other states in terms of
        accuracy and demographic parity violation. \textbf{Bottom:} Each panel
        shows an GBM with post-processing to enforce demographic parity
        on the state on which it was trained and evaluated both ID and OOD on
        all 50 states. Confidence intervals are 95\% Clopper-Pearson intervals for accuracy 
        and 95\% Newcombe intervals for demographic parity.
    }
    \label{fig:pumsadult_state_transfer}
\end{figure*}

Due to space constraints, we focus our experiments in this section on the
ACSIncome prediction task with demographic parity as the fairness criterion of
interest.  We present similar results for our other prediction tasks and
fairness criteria, as well as full experimental details in
Appendix~\ref{app:additional_experiments}.

\paragraph{Intervention effect sizes vary across states.}
The fifty US states which comprise the ACS PUMS data present a broad set of
different experimental conditions on which to evaluate the performance of
fairness interventions. At the most basic level, we can train and evaluate
different fairness interventions on each of the states and compare the
interventions' efficacy on these different distributions.  Concretely, we first
train an unconstrained gradient boosted decision tree (GBM) on each state, and
we compare the accuracy and fairness criterion violation of this unconstrained
model with the same model after applying one of three common fairness
intervention: pre-processing (LFR), the in-processing fair reductions methods
from~\citet{agarwal2018reductions} (ExpGrad), and the simple post-processing
method that adjusts group-based acceptance thresholds to satisfy a
constraint~\cite{hardt2016equality}.  Figure~\ref{fig:PUMSAdult_fairness_flow}
shows the result of this experiment for the ACSIncome prediction task for
interventions to achieve demographic parity.  For a given method, performance
can differ markedly between states. For instance, LFR decreases the demographic
parity violation by 10\% in some states and in other states the decrease is
close to zero.  Similarly, the post-processing adjustment to enforce demographic
parity incurs accuracy drops of less than 1\% in some states, whereas in others
the drop is closer to 5\%.

\paragraph{Training and testing on different states leads to unpredictable results.}
Beyond training and evaluating interventions on different states, we also use
the ACS PUMS data to study the performance of interventions under
\emph{geographic} distribution shift, where we train a model on one state and test it
on another. In Figure~\ref{fig:pumsadult_state_transfer}, we plot accuracy and
demographic parity violation with respect to race for both an unconstrained
GBM and the same model after applying a post-processing adjustment to achieve
demographic parity on a natural suite of test sets: the in-distribution (same state test set) and the
out-of-distribution test sets for the 49 other states.  For both the unconstrained and post-processed
model, model accuracy and demographic parity violation varies substantially across different state test sets. 
In particular, even when a method achieves demographic parity in one state, it
may no longer satisfy the fairness constraint when naively deployed on another.

\begin{figure*}
    \centering
    \includegraphics[width=\linewidth]{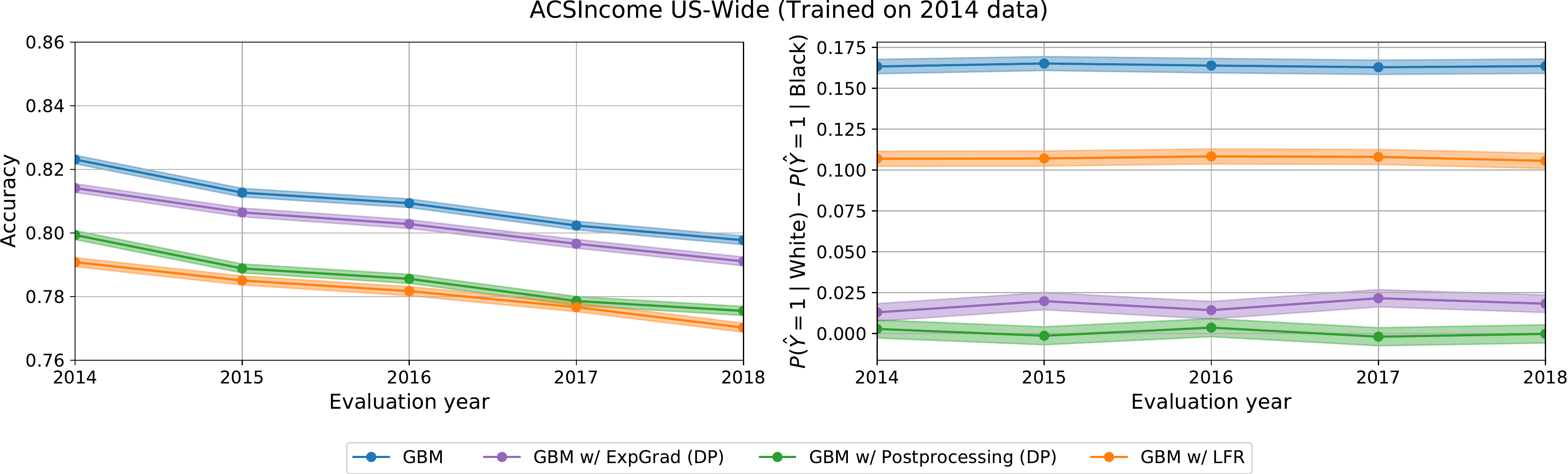}
    \caption{
        Fairness criteria are more stable over time than accuracy.
        \textbf{Left:} Models trained in 2014 on US-wide ACSIncome with and
        without fairness interventions to achieve demographic parity and
        evaluated on data in subsequent years suffer a drop in
        accuracy over time. \textbf{Right:} However, the violation of
        demographic parity remains essentially constant over time. Confidence
        intervals are 95\% Clopper-Pearson intervals for accuracy and 95\%
        Newcombe intervals for demographic parity.
    }
    \label{fig:pumsadult_time}
\end{figure*}

\paragraph{Fairness criteria are more stable over time than predictive
accuracy.}
In contrast to the unpredictable results that occur under geographic
distribution shift, the fairness criteria and interventions we study are much
more stable under \emph{temporal} distribution shift. 
Specifically, in Figure~\ref{fig:pumsadult_time}, we plot model accuracy and
demographic parity violation for GBM trained on the ACSIncome task using
US-wide data from 2014 and evaluated on the test sets for the same task drawn
from years 2014-2018. Perhaps unsurprisingly, model accuracy degrades slightly over
time. However, the associated fairness metric is stable and essentially constant
over time. Moreover, this same trend holds for the fairness interventions
previously discussed. The same base GBM with pre-processing (LFR),
in-processing (ExpGrad), or post-processing to satisfy demographic parity in
2014, all have a similar degradation in accuracy, but the fairness metrics remain stable. Thus, a classifier that satisfies demographic parity on the 2014 data
continues to satisfy the constraint on 2015-2018 data.

\vspace{-0.1cm}
\subsection{Specifying a locus of intervention}
On the ACSPUMs prediction task, fairness interventions can be applied either on
a state-by-state basis or on the entire US population. In
Table~\ref{table:intervention_locus}, we compare the performance of LFR and the
post-processing adjustment method applied at the US-level with the aggregate
performance of both methods applied on a state-by-state basis, using a GBM as
the base classifier. In both cases, applying the intervention on a state-by-state
improves US-wide accuracy while still preserving demographic parity
(post-processing) or further mitigating violations of demographic parity (LFR).

\begin{table}
    \caption{Comparison of two different strategies for applying an intervention
    to achieve demographic parity (DP) on the US-wide ACSIncome task.
    \emph{US-level} corresponds to training one classifier and applying the
    intervention on the entire US population. \emph{State-level} corresponds to
    training a classifier and applying the intervention separately for each
    state and then aggregating the results over all states. Here, DP
    refers to $P(\hat{Y} = 1\mid \mathrm{White}) - P(\hat{Y} = 1 \mid \mathrm{Black})$. 
    Confidence intervals are 95\% Clopper-Pearson intervals for accuracy and 95\% Newcombe intervals for
    DP. }
  \label{table:intervention_locus}
  \centering
  \begin{tabular}{lccccc}
    \toprule
        & US-level acc & \makecell{US-level \\ DP violation} & State-level acc & \makecell{State-level\\ DP violation}\\
    \midrule
    Unconstrained GBM &  $81.7 \pm 0.1$ \%& $17.7 \pm 0.2$\% & $82.8 \pm 0.1$ \% & $16.9 \pm 0.2$\%\\
     GBM w/ LFR &  $78.7 \pm 0.1$ \% & $16.6 \pm 0.2$\% & $79.4 \pm 0.1$\% & $14.0 \pm 0.2$\%\\
    GBM w/ post-processing (DP) & $79.2 \pm 0.1$ \% & $0.3 \pm 0.3$ \%& $80.2 \pm 0.1$\% & $-0.6 \pm 0.3$\%\\
    \bottomrule
  \end{tabular}
\end{table}

\subsection{Increased dataset size doesn't necessarily mitigate observed disparities}
To mitigate disparities in error rates, commonly suggested remedies include collecting a) larger datasets and b) more representative data reflective of social progress. For example, in response to research revealing the stark accuracy disparities of commercial facial recognition algorithms, particularly for dark-skinned females \citep{buolamwini2018gender}, IBM collected a more diverse training set of images, retrained its facial recognition model, and reported a 10-fold decrease in error for this subgroup \citep{puri_2019}. 
However, on our tabular datasets, larger datasets collected in more
socially progressive times do not automatically mitigate disparities. Table
\ref{table:persistent_error} shows that unconstrained gradient boosted decision
tree trained on a newer, larger dataset (ACSIncome vs. IPUMS Adult), does not
improve disparities such as in true positive rate (TPR). A fundamental reason
for this is the persistent social inequality that is reflected in the data. It
is well known that given a disparity in base rates between groups, a predictive
model cannot be both calibrated and  equal in error rates across groups
\citep{chouldechova2017fair}, except if the model has 100\% accuracy. This
observation highlights a key difference between cognitive machine learning and
tabular data prediction -- the Bayes error rate is zero for cognitive machine
learning. Thus larger and more representative datasets eventually address
disparities by pushing error rates to zero for all subgroups. In the tabular
datasets we collect, the Bayes error rate of an optimal classifier is almost
certainly far from zero, so some individuals will inevitably be incorrectly
classified. Rather than hope for future datasets to implicitly address
disparities, we must directly contend with how dataset and model design choices
distribute the burden of these errors.

\begin{table}
  \caption{Disparities persist despite increasing dataset size and social progress.}
  \label{table:persistent_error}
  \centering
  \begin{tabular}{lcccccc}
    \toprule
    Dataset & Year & Datapoints & GBM acc & TPR White & TPR Black & TPR disparity \\
    \midrule
    IPUMS Adult & 1994 & 49,531 & 86.4\% & 58.0\% & 46.5 \% & 11.5\%\\
    ACSIncome & 2018 & 1,599,229 & 80.8\% & 66.5\% & 51.7\% & 14.8\%\\
    \bottomrule
  \end{tabular}
\end{table}

\section{Discussion and future directions}
\label{sec:discussion}

Rather than settled conclusions, our empirical observations are intended to spark additional work on our new datasets. Of particular interest is a broad and comprehensive evaluation of existing methods on all datasets. We only evaluated some methods so far. One interesting question is if there is a method for achieving either demographic parity or error rate parity that outperforms threshold adjustment (based on the best known unconstrained classifier) on any of our datasets? We conjecture that the answer is \emph{no}. The reason is that we believe on our datasets a well-tuned tree-ensemble achieves classification error close to the Bayes error bound. Existing theory (Theorem 5.3 in \cite{hardt2016equality}) would then show that threshold adjustment based on this model is, in fact, optimal. Our conjecture motivates drawing a distinction between classification scenarios where a nearly Bayes optimal classifier is known and those where there isn't. How close we are to Bayes optimal on any of our new prediction tasks is a good question.
The role of distribution shift also deserves more attention. Are there methods that achieve consistent performance across geographic contexts? Why does there appear to be more temporal than geographic stability? What does the sensitivity to distribution shift say about algorithmic tools developed in one context and deployed in another? Answers to these questions seem highly relevant to policy-making around the deployment of algorithmic risk assessment tools.
Finally, our datasets are also interesting test cases for causal inference methods, which we haven't yet explored. How would, for example, methods like \emph{invariant risk minimization} \citep{arjovsky2019invariant} perform on different geographic contexts?

\section*{Acknowledgements}
We thank Barry Becker and Ronny Kohavi for answering our many questions around
the origin and creation of the UCI Adult dataset.  FD and JM are supported by
the National Science Foundation Graduate Research Fellowship Program under Grant
No. DGE 1752814. FD is additionally supported by the Open Philanthropy Project
AI Fellows Program.

\bibliographystyle{abbrvnat}
\bibliography{pums}

\clearpage
\newpage
\appendix

\etocdepthtag.toc{mtappendix}
\etocsettagdepth{mtsection}{none}
\etocsettagdepth{mtappendix}{subsubsection}
\tableofcontents

\section{Adult reconstruction}
\label{app:adult_reconstruction}

\subsection{Additional reconstruction details}
We only included a candidate variable if the nearest neighbor match was
\emph{exact}, i.e., we could find an exact match in the IPUMS CPS data for each
row in UCI Adult that matched \emph{both} the candidate variable and all earlier
variables also identified via exact matches. There were only two exceptions to
this rule:
\begin{itemize}
\item The UCI Adult feature ``native-country''. Here we could match the vast majority of rows in UCI Adult to the IPUMS CPS variable ``UH\_NATVTY\_A1''. To get an exact match for all rows, we had to map the country codes for Russia and Guyana in ``UH\_NATVTY\_A1'' to the value for ``unknown''. The documentation for UCI Adult also mentions neither Russia nor Guyana as possible values for ``native-country''. We do not know the reason for this discrepancy.
\item The UCI Adult feature ``fnlwgt''. This column is actually not a demographic feature of an individual but a weight value computed by the Census Bureau to make the sample representative for the US population.
We compared the ``fnlwgt'' data to all weight variables available in IPUMS CPS but did not find an exact match.
The closest match is the variable ``UH\_WGTS\_A1'', which has a similar distribution.
Since we did not identify an exact match for ``fnlwgt'' and the variable is not a property of an individual, we do not utilize it further in our experiments.
\end{itemize}

\subsection{Varying the income threshold experiments}

In our experiments, we randomly split the 49,531 examples in the IPUMS Adult
reconstruction into a training set of size 32,094 and a test-set of size 13,755.
We vary the threshold from 6,000 to 72,000. Concretely, for a given threshold,
e.g. 25,000, the task is to predict whether the individual's income is
greater than 25,000. We use a one-hot encoding for the categorical features, and
we use the same clustering preprocessing for the {\tt Education-Num} and {\tt
Age} features as~\citet{bellamy2019ai}. All features are further scaled to be
zero-mean and have unit variance.

In our experiments, as the ``unconstrained'' base classifier, we use the
gradient boosted decision tree classifier provided by~\citet{scikit_learn} with
exponential loss, {\tt num\_estimators} 5, {\tt max\_depth} 5, and all other
hyperparameters set to the default. We found this to slightly outperform the
default gradient boosting machine at threshold 50,000. For the three fairness
interventions, we used the implementation of LFR~\cite{zemel2013learning}
provided by~\citet{bellamy2019ai} with hyperparameters {\tt Ax} 1e-4, {\tt Ay}
1.0, {\tt Az} 1000, {\tt maxiter} 20000, and {\tt maxfun} 20000, which were
chosen by a grid search at threshold 50,000 to maximize the difference between
accuracy and the demographic parity disparity. We used the implementation of the
reductions approach of~\citet{agarwal2018reductions} provided
by~\citet{bird2020fairlearn} with the default hyperparameters, and we used
implementation of post-processing~\citep{hardt2016equality} provided
by~\citet{bellamy2019ai}.

\begin{figure*}
    \centering
    \includegraphics[width=\linewidth]{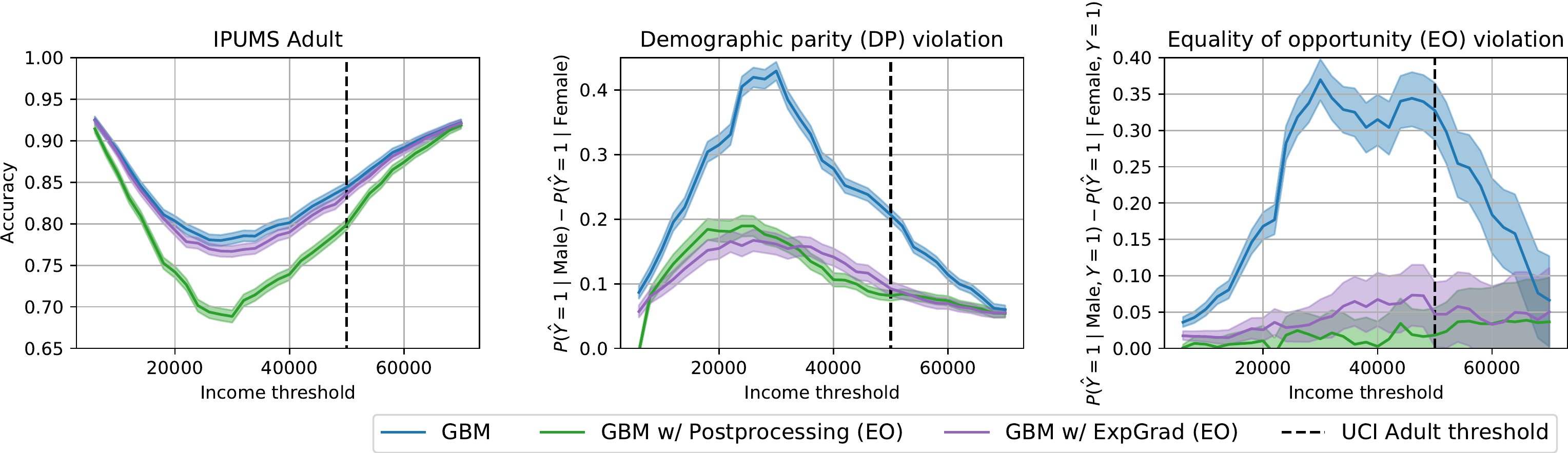}
    \caption{
        Fairness interventions with varying income threshold on IPUMS Adult.
        Comparison of in-processing and post-processing methods for achieving
        equality of opportunity (EO). LFR does not target EO, so we exclude it
        from the comparison.  Confidence intervals are 95\% Clopper-Pearson
        intervals for accuracy and 95\% Newcombe intervals for equality of
        opportunity.
    }
    \label{fig:adult_vary_threshold_eo}
\end{figure*}

In Figure~\ref{fig:adult_vary_threshold} in the main text, we compare the
performance of these three fairness interventions when enforcing demographic
parity as the threshold varies.  In Figure~\ref{fig:adult_vary_threshold_eo}, we
additionally compare the performance of in-processing method (ExpGrad) and the
post-processing method when enforcing equality of opportunity (EO). We exclude
LFR from the comparison because this method does not enforce equality of
opportunity without additional modification. The results from this experiment
are very similar to the experiment enforcing demographic parity. As the
threshold varies, the accuracy drop needed to enforce EO varies substantially,
as does the trade-off between criteria when enforcing EO. Moreover, for high
values of the threshold, the small number of positive instances substantially
increases the confidence intervals around the report EO values and makes it
difficult to compare the different interventions.

\section{New prediction task details}
\label{appendix:new-tasks}
In this section we detail the target variable, features, and filters that comprise each of our prediction tasks; more information about each feature can be found from the ACS PUMS documentation.\footnote{https://www.census.gov/programs-surveys/acs/microdata/documentation.html} For each feature, we list the variable code as provided by the ACS PUMS data sample, its extended description in parentheses, and finally the range of values for the variable.

\subsection{ACSIncome} 
Predict whether US working adults' yearly income is above \$50,000.

\paragraph{Target:} PINCP (Total person's income): an individual's label is 1 if PINCP > 50000, otherwise 0. Note that with our software package, this chosen income threshold can be toggled easily to label the ACS PUMS data differently, and construct a new prediction task.

\paragraph{Features:}
\begin{itemize}
	\item AGEP (Age): Range of values:
	\begin{itemize}
		\item  0 - 99 (integers)
		\item 0 indicates less than 1 year old. 
	\end{itemize}
	\item COW (Class of worker): Range of values:
		\begin{itemize}
			\item N/A (not in universe)
			\item 1: Employee of a private for-profit company or business, or of an individual, for wages, salary, or commissions
			\item 2: Employee of a private not-for-profit, tax-exempt, or charitable organization
			\item 3: Local government employee (city, county, etc.)
			\item 4: State government employee
			\item 5: Federal government employee
			\item 6: Self-employed in own not incorporated business, professional practice, or farm
			\item 7: Self-employed in own incorporated business, professional practice or farm
			\item 8: Working without pay in family business or farm
			\item 9: Unemployed and last worked 5 years ago or earlier or never worked 
	\end{itemize}
	\item SCHL (Educational attainment): Range of values: 
		\begin{itemize}
		\item N/A (less than 3 years old)
		\item 1: No schooling completed
		\item 2: Nursery school/preschool
		\item 3: Kindergarten
		\item 4: Grade 1
		\item 5: Grade 2
		\item 6: Grade 3
		\item 7: Grade 4
		\item 8: Grade 5
		\item 9: Grade 6
		\item 10: Grade 7
		\item 11: Grade 8
		\item 12: Grade 9
		\item 13: Grade 10
		\item 14: Grade 11
		\item 15: 12th Grade - no diploma
		\item 16: Regular high school diploma
		\item 17: GED or alternative credential
		\item 18: Some college but less than 1 year
		\item 19: 1 or more years of college credit but no degree
		\item 20: Associate's degree
		\item 21: Bachelor's degree
		\item 22: Master's degree
		\item 23: Professional degree beyond a bachelor's degree
		\item 24: Doctorate degree
		\end{itemize}

	\item MAR (Marital status): Range of values: 
	\begin{itemize}
	\item 1: Married
	\item 2: Widowed 
	\item 3: Divorced
	\item 4: Separated
	\item 5: Never married or under 15 years old
	\end{itemize}
	
	\item OCCP (Occupation): Please see ACS PUMS documentation for the full list of occupation codes
	\item POBP (Place of birth): Range of values includes most countries and individual US states; please see ACS PUMS documentation for the full list.
	\item RELP (Relationship): Range of values: 
		\begin{itemize}
		\item 0:	 Reference person
		\item 1:	Husband/wife
		\item 2:	Biological son or daughter
		\item 3:	Adopted son or daughter
		\item 4:	Stepson or stepdaughter
		\item 5:	Brother or sister
		\item 6:	Father or mother
		\item 7:	Grandchild
		\item 8:	Parent-in-law
		\item 9:	Son-in-law or daughter-in-law
		\item 10:	Other relative
		\item 11:	Roomer or boarder
		\item 12:	Housemate or roommate
		\item 13:	Unmarried partner
		\item 14:	Foster child
		\item 15:	Other nonrelative
		\item 16:	Institutionalized group quarters population
		\item 17:	Noninstitutionalized group quarters population
		\end{itemize}

	\item WKHP (Usual hours worked per week past 12 months): Range of values: 
		\begin{itemize}
		\item N/A (less than 16 years old / did not work during the past 12 months)
		\item 1 - 98 integer valued: usual hours worked
		\item 99: 99 or more usual hours
		\end{itemize}
	\item SEX (Sex): Range of values: 
		\begin{itemize}
		\item 1: Male
		\item 2: Female
		\end{itemize}
	\item RAC1P (Recoded detailed race code): Range of values: 
		\begin{itemize}
		\item 1:	White alone
		\item 2:	Black or African American alone
		\item 3:	American Indian alone
		\item 4:	Alaska Native alone
		\item 5:	American Indian and Alaska Native tribes specified, or American Indian or Alaska Native, not specified and no other races
		\item 6:	Asian alone
		\item 7:	Native Hawaiian and Other Pacific Islander alone
		\item 8:	Some Other Race alone
		\item 9:	Two or More Races
		\end{itemize}
\end{itemize}

\paragraph{Filters:}
\begin{itemize}
	\item AGEP (Age): Must be greater than 16
	\item PINCP (Total person's income): Must be greater than 100
	\item WKHP (Usual hours worked per week past 12 months): Must be greater than 0
	\item PWGTP (Person weight (relevant for re-weighting dataset to represent the general US population most accurately)): Must be greater than or equal to 1
\end{itemize}

\subsection{ACSPublicCoverage}
Predict whether a low-income individual, not eligible for Medicare, has coverage from public health insurance.

\paragraph{Target:} PUBCOV (Public health coverage): an individual's label is 1 if PUBCOV == 1 (with public health coverage), otherwise 0.

\paragraph{Features:}
\begin{itemize}
	\item AGEP (Age): Range of values:
	\begin{itemize}
		\item 0 - 99 (integers)
		\item 0 indicates less than 1 year old. 
	\end{itemize}

	\item SCHL (Educational attainment): Range of values: 
		\begin{itemize}
		\item N/A (less than 3 years old)
		\item 1: No schooling completed
		\item 2: Nursery school/preschool
		\item 3: Kindergarten
		\item 4: Grade 1
		\item 5: Grade 2
		\item 6: Grade 3
		\item 7: Grade 4
		\item 8: Grade 5
		\item 9: Grade 6
		\item 10: Grade 7
		\item 11: Grade 8
		\item 12: Grade 9
		\item 13: Grade 10
		\item 14: Grade 11
		\item 15: 12th Grade - no diploma
		\item 16: Regular high school diploma
		\item 17: GED or alternative credential
		\item 18: Some college but less than 1 year
		\item 19: 1 or more years of college credit but no degree
		\item 20: Associate's degree
		\item 21: Bachelor's degree
		\item 22: Master's degree
		\item 23: Professional degree beyond a bachelor's degree
		\item 24: Doctorate degree
		\end{itemize}

	\item MAR (Marital status): Range of values: 
	\begin{itemize}
	\item 1: Married
	\item 2: Widowed 
	\item 3: Divorced
	\item 4: Separated
	\item 5: Never married or under 15 years old
	\end{itemize}
	
	\item SEX (Sex): Range of values: 
		\begin{itemize}
		\item 1: Male
		\item 2: Female
		\end{itemize}
		
	\item DIS (Disability recode): Range of values: 
	\begin{itemize}
	\item 1: With a disability
	\item 2: Without a disability
	\end{itemize}
	
	\item ESP (Employment status of parents): Range of values:
	\begin{itemize}
	\item N/A (not own child of householder, and not child in subfamily)
	\item 1:	Living with two parents: both parents in labor force
	\item 2:	Living with two parents: Father only in labor force
	\item 3:	Living with two parents: Mother only in labor force
	\item 4:	Living with two parents: Neither parent in labor force
	\item 5:	Living with father: Father in the labor force
	\item 6:	Living with father: Father not in labor force
	\item 7:	Living with mother: Mother in the labor force
	\item 8:	Living with mother: Mother not in labor force
	\end{itemize}

	\item CIT (Citizenship status): Range of values:
	\begin{itemize}
	\item 1:	Born in the U.S.
	\item 2:	Born in Puerto Rico, Guam, the U.S. Virgin Islands, or the Northern Marianas
	\item 3:	Born abroad of American parent(s)
	\item 4:	U.S. citizen by naturalization
	\item 5:	Not a citizen of the U.S.
	\end{itemize}

	\item MIG (Mobility status (lived here 1 year ago): Range of values: 
	\begin{itemize}
	\item N/A (less than 1 year old)
	\item 1:	Yes, same house (nonmovers)
	\item 2:	No, outside US and Puerto Rico
	\item 3:	No, different house in US or Puerto Rico
	\end{itemize}

	\item MIL (Military service): Range of values: 
	\begin{itemize}
	\item N/A (less than 17 years old)
	\item 1:	Now on active duty
	\item 2:	On active duty in the past, but not now
	\item 3:	Only on active duty for training in Reserves/National Guard
	\item 4:	Never served in the military
	\end{itemize}

	\item ANC (Ancestry recode): Range of values:
	\begin{itemize}
	\item 1:	Single
	\item 2:	Multiple
	\item 3:	Unclassified
	\item 4:	Not reported
	\item 8:	Suppressed for data year 2018 for select PUMAs
	\end{itemize}

	\item NATIVITY (Nativity): Range of values: 
	\begin{itemize}
	\item 1: Native
	\item 2: Foreign born
	\end{itemize}

	\item DEAR (Hearing difficulty): Range of values: 
	\begin{itemize}
	\item 1: Yes
	\item 2: No
	\end{itemize}

	\item DEYE (Vision difficulty): Range of values: 
	\begin{itemize}
	\item 1: Yes
	\item 2: No
	\end{itemize}
	
	\item DREM (Cognitive difficulty): Range of values: 
	\begin{itemize}
	\item N/A (less than 5 years old)
	\item 1: Yes
	\item 2: No
	\end{itemize}

	\item PINCP (Total person's income): Range of values: 
	\begin{itemize}
	\item integers between -19997 and 4209995 to indicate income in US dollars
	\item loss of \$19998 or more is coded as -19998.
	\item income of \$4209995 or more is coded as 4209995.
	\end{itemize}
	
	\item ESR (Employment status recode): Range of values:
	\begin{itemize}
	\item N/A (less than 16 years old)
	\item 1:	Civilian employed, at work
	\item 2:	Civilian employed, with a job but not at work
	\item 3:	Unemployed
	\item 4:	Armed forces, at work
	\item 5:	Armed forces, with a job but not at work
	\item 6:	Not in labor force
	\end{itemize}

	\item ST (State code): Please see ACS PUMS documentation for the correspondence between coded values and state name.
	\item FER (Gave birth to child within the past 12 months): Range of values:
	\begin{itemize}
	\item N/A (less than 15 years/greater than 50 years/male)
	\item 1:	Yes
	\item 2:	No
	\end{itemize}
	
	\item RAC1P (Recoded detailed race code): Range of values: 
		\begin{itemize}
		\item 1:	White alone
		\item 2:	Black or African American alone
		\item 3:	American Indian alone
		\item 4:	Alaska Native alone
		\item 5:	American Indian and Alaska Native tribes specified, or American Indian or Alaska Native, not specified and no other races
		\item 6:	Asian alone
		\item 7:	Native Hawaiian and Other Pacific Islander alone
		\item 8:	Some Other Race alone
		\item 9:	Two or More Races
		\end{itemize}
		
\end{itemize}

\paragraph{Filters:}
\begin{itemize}
	\item AGEP (Age) must be less than 65.
	\item PINCP (Total person's income) must be less than \$30,000.
\end{itemize}

\subsection{ACSMobility} 
Predict whether a young adult moved addresses in the last year.

\paragraph{Target:} MIG (Mobility status): an individual's label is 1 if MIG == 1, and 0 otherwise.

\paragraph{Features:}
\begin{itemize}
	\item AGEP (Age): Range of values:
	\begin{itemize}
		\item 0 - 99 (integers)
		\item 0 indicates less than 1 year old. 
	\end{itemize}

	\item SCHL (Educational attainment): Range of values: 
		\begin{itemize}
		\item N/A (less than 3 years old)
		\item 1: No schooling completed
		\item 2: Nursery school/preschool
		\item 3: Kindergarten
		\item 4: Grade 1
		\item 5: Grade 2
		\item 6: Grade 3
		\item 7: Grade 4
		\item 8: Grade 5
		\item 9: Grade 6
		\item 10: Grade 7
		\item 11: Grade 8
		\item 12: Grade 9
		\item 13: Grade 10
		\item 14: Grade 11
		\item 15: 12th Grade - no diploma
		\item 16: Regular high school diploma
		\item 17: GED or alternative credential
		\item 18: Some college but less than 1 year
		\item 19: 1 or more years of college credit but no degree
		\item 20: Associate's degree
		\item 21: Bachelor's degree
		\item 22: Master's degree
		\item 23: Professional degree beyond a bachelor's degree
		\item 24: Doctorate degree
		\end{itemize}

	\item MAR (Marital status): Range of values: 
	\begin{itemize}
	\item 1: Married
	\item 2: Widowed 
	\item 3: Divorced
	\item 4: Separated
	\item 5: Never married or under 15 years old
	\end{itemize}
	
	\item SEX (Sex): Range of values: 
		\begin{itemize}
		\item 1: Male
		\item 2: Female
		\end{itemize}
		
	\item DIS (Disability recode): Range of values: 
	\begin{itemize}
	\item 1: With a disability
	\item 2: Without a disability
	\end{itemize}
	
	\item ESP (Employment status of parents): Range of values:
	\begin{itemize}
	\item N/A (not own child of householder, and not child in subfamily)
	\item 1:	Living with two parents: both parents in labor force
	\item 2:	Living with two parents: Father only in labor force
	\item 3:	Living with two parents: Mother only in labor force
	\item 4:	Living with two parents: Neither parent in labor force
	\item 5:	Living with father: Father in the labor force
	\item 6:	Living with father: Father not in labor force
	\item 7:	Living with mother: Mother in the labor force
	\item 8:	Living with mother: Mother not in labor force
	\end{itemize}

	\item CIT (Citizenship status): Range of values:
	\begin{itemize}
	\item 1:	Born in the U.S.
	\item 2:	Born in Puerto Rico, Guam, the U.S. Virgin Islands, or the Northern Marianas
	\item 3:	Born abroad of American parent(s)
	\item 4:	U.S. citizen by naturalization
	\item 5:	Not a citizen of the U.S.
	\end{itemize}

	\item MIL (Military service): Range of values: 
	\begin{itemize}
	\item N/A (less than 17 years old)
	\item 1:	Now on active duty
	\item 2:	On active duty in the past, but not now
	\item 3:	Only on active duty for training in Reserves/National Guard
	\item 4:	Never served in the military
	\end{itemize}

	\item ANC (Ancestry recode): Range of values:
	\begin{itemize}
	\item 1:	Single
	\item 2:	Multiple
	\item 3:	Unclassified
	\item 4:	Not reported
	\item 8:	Suppressed for data year 2018 for select PUMAs
	\end{itemize}

	\item NATIVITY (Nativity): Range of values: 
	\begin{itemize}
	\item 1: Native
	\item 2: Foreign born
	\end{itemize}
	
	\item RELP (Relationship): Range of values: 
		\begin{itemize}
		\item 0:	 Reference person
		\item 1:	Husband/wife
		\item 2:	Biological son or daughter
		\item 3:	Adopted son or daughter
		\item 4:	Stepson or stepdaughter
		\item 5:	Brother or sister
		\item 6:	Father or mother
		\item 7:	Grandchild
		\item 8:	Parent-in-law
		\item 9:	Son-in-law or daughter-in-law
		\item 10:	Other relative
		\item 11:	Roomer or boarder
		\item 12:	Housemate or roommate
		\item 13:	Unmarried partner
		\item 14:	Foster child
		\item 15:	Other nonrelative
		\item 16:	Institutionalized group quarters population
		\item 17:	Noninstitutionalized group quarters population
		\end{itemize}

	\item DEAR (Hearing difficulty): Range of values: 
	\begin{itemize}
	\item 1: Yes
	\item 2: No
	\end{itemize}

	\item DEYE (Vision difficulty): Range of values: 
	\begin{itemize}
	\item 1: Yes
	\item 2: No
	\end{itemize}
	
	\item DREM (Cognitive difficulty): Range of values: 
	\begin{itemize}
	\item N/A (less than 5 years old)
	\item 1: Yes
	\item 2: No
	\end{itemize}

	\item RAC1P (Recoded detailed race code): Range of values: 
		\begin{itemize}
		\item 1:	White alone
		\item 2:	Black or African American alone
		\item 3:	American Indian alone
		\item 4:	Alaska Native alone
		\item 5:	American Indian and Alaska Native tribes specified, or American Indian or Alaska Native, not specified and no other races
		\item 6:	Asian alone
		\item 7:	Native Hawaiian and Other Pacific Islander alone
		\item 8:	Some Other Race alone
		\item 9:	Two or More Races
		\end{itemize}

	\item GCL (Grandparents living with grandchildren): Range of values: 
	\begin{itemize}
		\item N/A (less than 30 years/institutional GQ)
		\item 1: Yes
		\item 2: No
	\end{itemize}

	\item COW (Class of worker): Range of values:
		\begin{itemize}
			\item N/A (not in universe)
			\item 1: Employee of a private for-profit company or business, or of an individual, for wages, salary, or commissions
			\item 2: Employee of a private not-for-profit, tax-exempt, or charitable organization
			\item 3: Local government employee (city, county, etc.)
			\item 4: State government employee
			\item 5: Federal government employee
			\item 6: Self-employed in own not incorporated business, professional practice, or farm
			\item 7: Self-employed in own incorporated business, professional practice or farm
			\item 8: Working without pay in family business or farm
			\item 9: Unemployed and last worked 5 years ago or earlier or never worked 
	\end{itemize}

	\item ESR (Employment status recode): Range of values:
	\begin{itemize}
	\item N/A (less than 16 years old)
	\item 1:	Civilian employed, at work
	\item 2:	Civilian employed, with a job but not at work
	\item 3:	Unemployed
	\item 4:	Armed forces, at work
	\item 5:	Armed forces, with a job but not at work
	\item 6:	Not in labor force
	\end{itemize}

	\item WKHP (Usual hours worked per week past 12 months): Range of values: 
			\begin{itemize}
			\item N/A (less than 16 years old / did not work during the past 12 months)
			\item 1 - 98 integer valued: usual hours worked
			\item 99: 99 or more usual hours
			\end{itemize}
	\item JWMNP (Travel time to work): Range of values: 
	\begin{itemize}
		\item N/A (not a worker  or a worker that worked at home)
		\item integers 1 - 200 for minutes to get to work 
		\item top-coded at 200 so values above 200 are coded as 200

	\end{itemize}
		\item PINCP (Total person's income): Range of values: 
	\begin{itemize}
	\item integers between -19997 and 4209995 to indicate income in US dollars
	\item loss of \$19998 or more is coded as -19998.
	\item income of \$4209995 or more is coded as 4209995.
	\end{itemize}

\end{itemize}

\paragraph{Filters:}
\begin{itemize}
	\item AGEP (Age) must be greater than 18 and less than 35.
\end{itemize}

\subsection{ACSEmployment}
Predict whether an adult is employed.

\paragraph{Target:} ESR (Employment status recode): an individual's label is 1 if ESR == 1, and 0 otherwise.

\paragraph{Features:}
\begin{itemize}
	\item AGEP (Age): Range of values:
	\begin{itemize}
		\item 0 - 99 (integers)
		\item 0 indicates less than 1 year old. 
	\end{itemize}

	\item SCHL (Educational attainment): Range of values: 
		\begin{itemize}
		\item N/A (less than 3 years old)
		\item 1: No schooling completed
		\item 2: Nursery school/preschool
		\item 3: Kindergarten
		\item 4: Grade 1
		\item 5: Grade 2
		\item 6: Grade 3
		\item 7: Grade 4
		\item 8: Grade 5
		\item 9: Grade 6
		\item 10: Grade 7
		\item 11: Grade 8
		\item 12: Grade 9
		\item 13: Grade 10
		\item 14: Grade 11
		\item 15: 12th Grade - no diploma
		\item 16: Regular high school diploma
		\item 17: GED or alternative credential
		\item 18: Some college but less than 1 year
		\item 19: 1 or more years of college credit but no degree
		\item 20: Associate's degree
		\item 21: Bachelor's degree
		\item 22: Master's degree
		\item 23: Professional degree beyond a bachelor's degree
		\item 24: Doctorate degree
		\end{itemize}

	\item MAR (Marital status): Range of values: 
	\begin{itemize}
	\item 1: Married
	\item 2: Widowed 
	\item 3: Divorced
	\item 4: Separated
	\item 5: Never married or under 15 years old
	\end{itemize}
	
	\item SEX (Sex): Range of values: 
		\begin{itemize}
		\item 1: Male
		\item 2: Female
		\end{itemize}
		
	\item DIS (Disability recode): Range of values: 
	\begin{itemize}
	\item 1: With a disability
	\item 2: Without a disability
	\end{itemize}
	
	\item ESP (Employment status of parents): Range of values:
	\begin{itemize}
	\item N/A (not own child of householder, and not child in subfamily)
	\item 1:	Living with two parents: both parents in labor force
	\item 2:	Living with two parents: Father only in labor force
	\item 3:	Living with two parents: Mother only in labor force
	\item 4:	Living with two parents: Neither parent in labor force
	\item 5:	Living with father: Father in the labor force
	\item 6:	Living with father: Father not in labor force
	\item 7:	Living with mother: Mother in the labor force
	\item 8:	Living with mother: Mother not in labor force
	\end{itemize}

	\item MIG (Mobility status (lived here 1 year ago): Range of values: 
	\begin{itemize}
	\item N/A (less than 1 year old)
	\item 1:	Yes, same house (nonmovers)
	\item 2:	No, outside US and Puerto Rico
	\item 3:	No, different house in US or Puerto Rico
	\end{itemize}
	
	\item CIT (Citizenship status): Range of values:
	\begin{itemize}
	\item 1:	Born in the U.S.
	\item 2:	Born in Puerto Rico, Guam, the U.S. Virgin Islands, or the Northern Marianas
	\item 3:	Born abroad of American parent(s)
	\item 4:	U.S. citizen by naturalization
	\item 5:	Not a citizen of the U.S.
	\end{itemize}
	
	\item MIL (Military service): Range of values: 
	\begin{itemize}
	\item N/A (less than 17 years old)
	\item 1:	Now on active duty
	\item 2:	On active duty in the past, but not now
	\item 3:	Only on active duty for training in Reserves/National Guard
	\item 4:	Never served in the military
	\end{itemize}

	\item ANC (Ancestry recode): Range of values:
	\begin{itemize}
	\item 1:	Single
	\item 2:	Multiple
	\item 3:	Unclassified
	\item 4:	Not reported
	\item 8:	Suppressed for data year 2018 for select PUMAs
	\end{itemize}

	\item NATIVITY (Nativity): Range of values: 
	\begin{itemize}
	\item 1: Native
	\item 2: Foreign born
	\end{itemize}
	
	\item RELP (Relationship): Range of values: 
		\begin{itemize}
		\item 0:	 Reference person
		\item 1:	Husband/wife
		\item 2:	Biological son or daughter
		\item 3:	Adopted son or daughter
		\item 4:	Stepson or stepdaughter
		\item 5:	Brother or sister
		\item 6:	Father or mother
		\item 7:	Grandchild
		\item 8:	Parent-in-law
		\item 9:	Son-in-law or daughter-in-law
		\item 10:	Other relative
		\item 11:	Roomer or boarder
		\item 12:	Housemate or roommate
		\item 13:	Unmarried partner
		\item 14:	Foster child
		\item 15:	Other nonrelative
		\item 16:	Institutionalized group quarters population
		\item 17:	Noninstitutionalized group quarters population
		\end{itemize}

	\item DEAR (Hearing difficulty): Range of values: 
	\begin{itemize}
	\item 1: Yes
	\item 2: No
	\end{itemize}

	\item DEYE (Vision difficulty): Range of values: 
	\begin{itemize}
	\item 1: Yes
	\item 2: No
	\end{itemize}
	
	\item DREM (Cognitive difficulty): Range of values: 
	\begin{itemize}
	\item N/A (less than 5 years old)
	\item 1: Yes
	\item 2: No
	\end{itemize}

	\item RAC1P (Recoded detailed race code): Range of values: 
		\begin{itemize}
		\item 1:	White alone
		\item 2:	Black or African American alone
		\item 3:	American Indian alone
		\item 4:	Alaska Native alone
		\item 5:	American Indian and Alaska Native tribes specified, or American Indian or Alaska Native, not specified and no other races
		\item 6:	Asian alone
		\item 7:	Native Hawaiian and Other Pacific Islander alone
		\item 8:	Some Other Race alone
		\item 9:	Two or More Races
		\end{itemize}

	\item GCL (Grandparents living with grandchildren): Range of values: 
	\begin{itemize}
		\item N/A (less than 30 years/institutional GQ)
		\item 1: Yes
		\item 2: No
	\end{itemize}
	
\end{itemize}

\paragraph{Filters:}
\begin{itemize}
	\item AGEP (Age) must be greater than 16 and less than 90.
	\item PWGTP (Person weight) must be greater than or equal to 1.
\end{itemize}

\subsection{ACSTravelTime} 
Predict whether a working adult has a travel time to work of greater than 20 minutes. 

\paragraph{Target:} JWMNP (Travel time to work): an individual's label is 1 if JWMNP > 20, and 0 otherwise.

\paragraph{Features:}
\begin{itemize}
	\item AGEP (Age): Range of values:
	\begin{itemize}
		\item 0 - 99 (integers)
		\item 0 indicates less than 1 year old. 
	\end{itemize}

	\item SCHL (Educational attainment): Range of values: 
		\begin{itemize}
		\item N/A (less than 3 years old)
		\item 1: No schooling completed
		\item 2: Nursery school/preschool
		\item 3: Kindergarten
		\item 4: Grade 1
		\item 5: Grade 2
		\item 6: Grade 3
		\item 7: Grade 4
		\item 8: Grade 5
		\item 9: Grade 6
		\item 10: Grade 7
		\item 11: Grade 8
		\item 12: Grade 9
		\item 13: Grade 10
		\item 14: Grade 11
		\item 15: 12th Grade - no diploma
		\item 16: Regular high school diploma
		\item 17: GED or alternative credential
		\item 18: Some college but less than 1 year
		\item 19: 1 or more years of college credit but no degree
		\item 20: Associate's degree
		\item 21: Bachelor's degree
		\item 22: Master's degree
		\item 23: Professional degree beyond a bachelor's degree
		\item 24: Doctorate degree
		\end{itemize}

	\item MAR (Marital status): Range of values: 
	\begin{itemize}
	\item 1: Married
	\item 2: Widowed 
	\item 3: Divorced
	\item 4: Separated
	\item 5: Never married or under 15 years old
	\end{itemize}
	
	\item SEX (Sex): Range of values: 
		\begin{itemize}
		\item 1: Male
		\item 2: Female
		\end{itemize}
		
	\item DIS (Disability recode): Range of values: 
	\begin{itemize}
	\item 1: With a disability
	\item 2: Without a disability
	\end{itemize}
	
	\item ESP (Employment status of parents): Range of values:
	\begin{itemize}
	\item N/A (not own child of householder, and not child in subfamily)
	\item 1:	Living with two parents: both parents in labor force
	\item 2:	Living with two parents: Father only in labor force
	\item 3:	Living with two parents: Mother only in labor force
	\item 4:	Living with two parents: Neither parent in labor force
	\item 5:	Living with father: Father in the labor force
	\item 6:	Living with father: Father not in labor force
	\item 7:	Living with mother: Mother in the labor force
	\item 8:	Living with mother: Mother not in labor force
	\end{itemize}

	\item MIG (Mobility status (lived here 1 year ago): Range of values: 
	\begin{itemize}
	\item N/A (less than 1 year old)
	\item 1:	Yes, same house (nonmovers)
	\item 2:	No, outside US and Puerto Rico
	\item 3:	No, different house in US or Puerto Rico
	\end{itemize}

	\item RELP (Relationship): Range of values: 
		\begin{itemize}
		\item 0:	 Reference person
		\item 1:	Husband/wife
		\item 2:	Biological son or daughter
		\item 3:	Adopted son or daughter
		\item 4:	Stepson or stepdaughter
		\item 5:	Brother or sister
		\item 6:	Father or mother
		\item 7:	Grandchild
		\item 8:	Parent-in-law
		\item 9:	Son-in-law or daughter-in-law
		\item 10:	Other relative
		\item 11:	Roomer or boarder
		\item 12:	Housemate or roommate
		\item 13:	Unmarried partner
		\item 14:	Foster child
		\item 15:	Other nonrelative
		\item 16:	Institutionalized group quarters population
		\item 17:	Noninstitutionalized group quarters population
		\end{itemize}

	\item RAC1P (Recoded detailed race code): Range of values: 
		\begin{itemize}
		\item 1:	White alone
		\item 2:	Black or African American alone
		\item 3:	American Indian alone
		\item 4:	Alaska Native alone
		\item 5:	American Indian and Alaska Native tribes specified, or American Indian or Alaska Native, not specified and no other races
		\item 6:	Asian alone
		\item 7:	Native Hawaiian and Other Pacific Islander alone
		\item 8:	Some Other Race alone
		\item 9:	Two or More Races
		\end{itemize}	
	
	\item PUMA (Public use microdata area code (PUMA) based on 2010 Census definition (areas with population of 100,000 or more, use with ST for unique code)): Please see ACS PUMS documentation for details on the PUMA codes (which range from 100 to 70301)
	\item ST (State code): Please see ACS PUMS documentation for the correspondence between coded values and state name.

	\item CIT (Citizenship status): Range of values:
	\begin{itemize}
	\item 1:	Born in the U.S.
	\item 2:	Born in Puerto Rico, Guam, the U.S. Virgin Islands, or the Northern Marianas
	\item 3:	Born abroad of American parent(s)
	\item 4:	U.S. citizen by naturalization
	\item 5:	Not a citizen of the U.S.
	\end{itemize}

	\item OCCP (Occupation): Please see ACS PUMS documentation for the full list of occupation codes
	
	\item JWTR (Means of transportation to work): Range of values: 	
	
	\begin{itemize}
	\item N/A (not a worker--not in the labor force, including persons under 16 years, unemployed, employed, with a job but not at work, Armed Forces, with a job but not at work)
	\item 1:	Car, truck, or van
	\item 2:	Bus or trolley bus
	\item 3:	Streetcar or trolley car (carro publico in Puerto Rico)
	\item 4:	Subway or elevated
	\item 5:	Railroad
	\item 6:	Ferryboat
	\item 7:	Taxicab
	\item 8:	Motorcycle
	\item 9:	Bicycle
	\item 10:	Walked;
	\item 11: Worked at home
	\item 12:	Other method
	\end{itemize}
	
	\item POWPUMA (Place of work PUMA based on 2010 Census definitions): Please see ACS PUMS documentation for details on PUMA codes
	\item POVPIP (Income-to-poverty ratio recode): Range of values: 
	\begin{itemize}
	\item N/A
	\item integers 0-500
	\item 501 for 501 percent or more
	\end{itemize}
\end{itemize}

\paragraph{Filters:}
\begin{itemize}
	\item AGEP (Age) must be greater than 16.
	\item PWGTP (Person weight) must be greater than or equal to 1.
	\item ESR (Employment status recode) must be equal to 1 (employed).
\end{itemize}

\subsection{Dataset access and license}
We provide a flexible software package to download ACS PUMS data and construct
both the new prediction tasks discussed in Section~\ref{sec:new_data}, as well as
new tasks using ACS PUMS data products. The ACS PUMS data itself is governed by
the terms of service from the US Census Bureau. For more information, see
\url{https://www.census.gov/data/developers/about/terms-of-service.html}
Similarly, the IPUMS adult reconstruction is governed by the IPUMS terms of
use. For more information, see \url{https://ipums.org/about/terms}.

\subsection{Table~\ref{table:tasks} experiment details}
For each of the tasks listed in Table~\ref{table:tasks} (ACSIncome,
ACSPublicCoverage, ACSMobility, ACSEmployment, ACSTravelTime), we use the 1-year
2018 US-Wide ACS PUMS data. We use a maximum of 100,000 examples from each
state, and randomly subsample states that have more than 100,000 examples.  We
randomly split 80\% of the dataset into a training split and the remaining 20\%
into a test split. All features are standardized to be zero-mean and
unit-variance.  {\tt Constant Predictor} refers to the majority class baseline,
{\tt LogReg} refers to a logistic regression baseline, and {\tt GBM} refers to a
gradient boosted decision tree classifier. For each models, we use the
implementation provided by~\citet{scikit_learn} with the default
hyperparameters.

\section{Tour of empirical observations: missing experimental details}
\label{app:experimental_details}
\paragraph{Models and hyperparameters.}
All of the experiments in this section use the same unconstrained base model: a
gradient boosted decision tree (GBM). 
We chose this model because it trains quickly and consistently achieved higher accuracy than other baseline models we considered (logistic regression and random forests) in the unconstrained setting; experiments using other base models also produced qualitatively similar results, so we focus on GBM in this paper.
We use the implementation provided
by~\citet{scikit_learn} and use {\tt exponential} loss, {\tt num\_estimators} 5,
{\tt max\_depth} 5, and all other hyperparameters set to the default. These
hyperparameters were chosen via a small grid search to maximize accuracy on the
ACSIncome task. We use the implementation of LFR~\citep{zemel2013learning}
from~\citet{bellamy2019ai} with hyperparameters {\tt k}=10, {\tt Ax}=0.1, {\tt
Ay}=1.0, {\tt Az} = 2.0, {\tt maxiter}=5000, and {\tt maxfun}=5000. The
hyperparameters are the same as those used in the UCI Adult tutorial provided
by~\citet{bellamy2019ai}. For the in-processing method (ExpGrad)
from~\citet{agarwal2018reductions}, we use the implementation
from~\citet{bird2020fairlearn} with the default hyperparameters, and for the
post-processing method, we use the threshold adjustment method
of~\citet{hardt2016equality}, which is also implemented
in~\citet{bellamy2019ai}. In Section~\ref{sec:experiments}, we use all of the
methods to enforce demographic parity. We detail additional experiments
enforcing equality of opportunity in Appendix~\ref{app:additional_experiments}.

\paragraph{Datasets.}
Throughout this section, we use the ACSIncome task described in
Section~\ref{sec:new_data} and Appendix~\ref{appendix:new-tasks}.
With the exception of the distribution shift across time experiments, we use the
2018 1-Year ACS PUMS data. For each state, we randomly split 80\% of the dataset
into a training split and use the remaining 20\% as a test split. The US-Wide
dataset is constructed by combining these training and testing sets over all 50
states and Puerto Rico. For the distribution shift across time experiments, we
use the same procedure for the 2014-2017 1-Year ACS PUMS data.

\paragraph{Confidence intervals.}
To account for random variation in estimating model accuracies and violations of
demographic parity and equality of opportunity, we report each of these metrics
with appropriate confidence intervals. We report and plot accuracy numbers with
95\% Clopper-Pearson intervals. We report and plot violations of demographic
parity and equality of opportunity with 95\% Newcombe intervals for the
difference between two binomial proportions.

\paragraph{Compute environment.}
All of our experiments are run on CPUs on a cluster computer with 24 Intel Xeon
E7 CPUs and 300 GB of RAM.

\section{Additional experiments}
\label{app:additional_experiments}
In this section, we conduct the same set of experiments conducted in
Section~\ref{sec:experiments} on the 5 other prediction tasks we introduced in
Section~\ref{sec:new_data}. Throughout we keep the experimental details (models,
hyperparameters, etc) identical to those detailed in
Appendix~\ref{app:experimental_details}.

\subsection{Intervention effect sizes across states}
\begin{figure*}
    \centering
    \includegraphics[width=\linewidth]{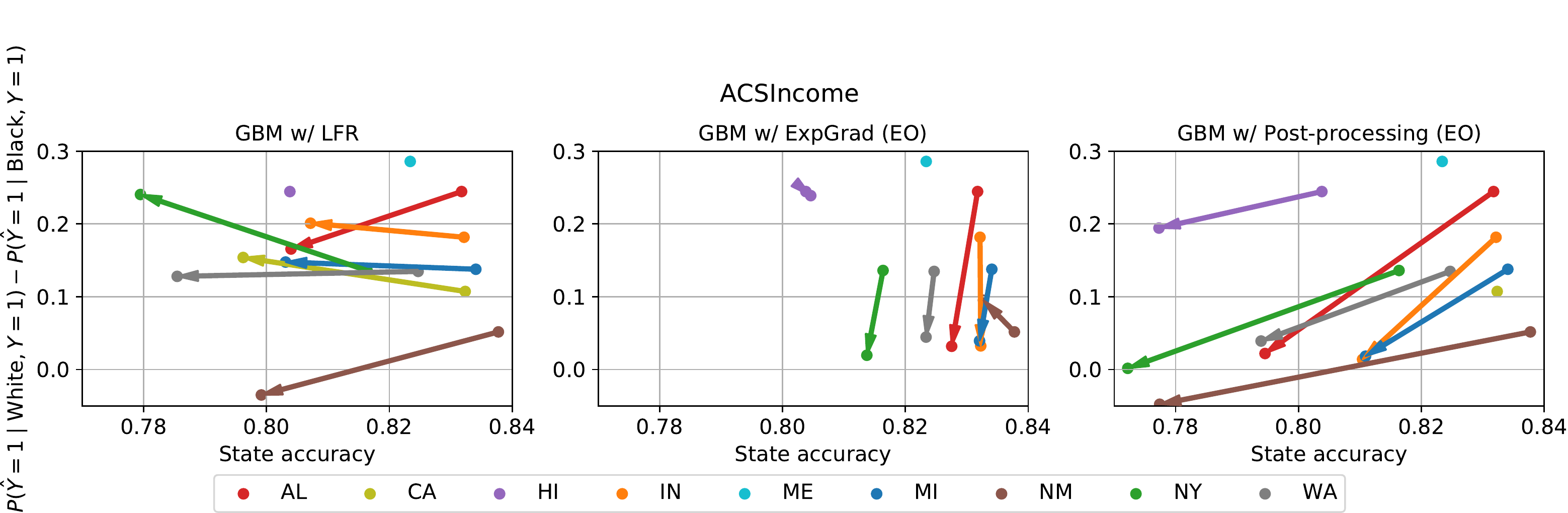}
    \caption{
        The effect size of fairness interventions varies by state. Each panel
        shows the change in accuracy and equality of opportunity violation (EO)
        on the ACSIncome task after applying a fairness intervention to an
        unconstrained gradient boosted decision tree (GBM). Each arrow
        corresponds to a different state distribution. The arrow base represents
        the (accuracy, EO) point corresponding to the unconstrained GBM, and the
        head represents the (accuracy, EO) point obtained after applying the
        intervention. The arrow for HI in the LFR plot and ME in all three plots
        is entirely covered by the start and end points.
    }
    \label{fig:PUMSAdult_fairness_flow_eo}
\end{figure*}

As in Section~\ref{sec:experiments}, we train an unconstrained gradient boosted
decision tree (GBM) on each state, and we compare the accuracy and fairness
criterion violation of this unconstrained model with the same model after
applying one of three common fairness intervention: pre-processing (LFR), the
in-processing fair reductions methods from~\citet{agarwal2018reductions}
(ExpGrad), and the simple post-processing method that adjusts group-based
acceptance thresholds to satisfy a constraint~\cite{hardt2016equality}.
Figure~\ref{fig:PUMSAdult_fairness_flow_eo} shows the result of this experiment for
the ACSIncome prediction task for interventions to achieve equality of
opportunity.

\begin{figure*}
    \centering
    \includegraphics[width=\linewidth]{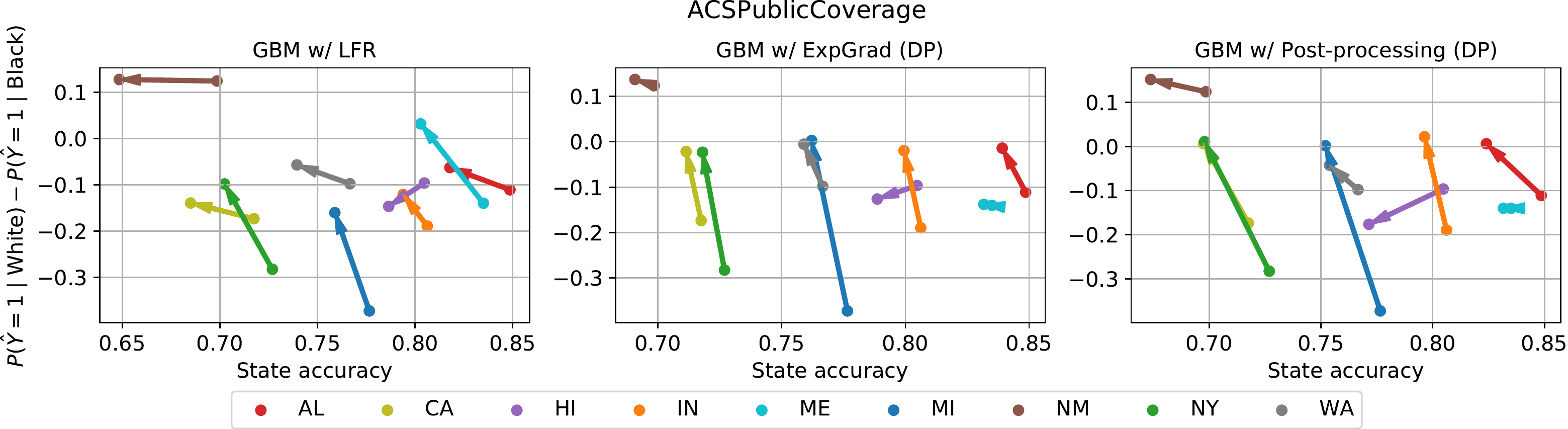}
    \vspace{0.2cm}
    \includegraphics[width=\linewidth]{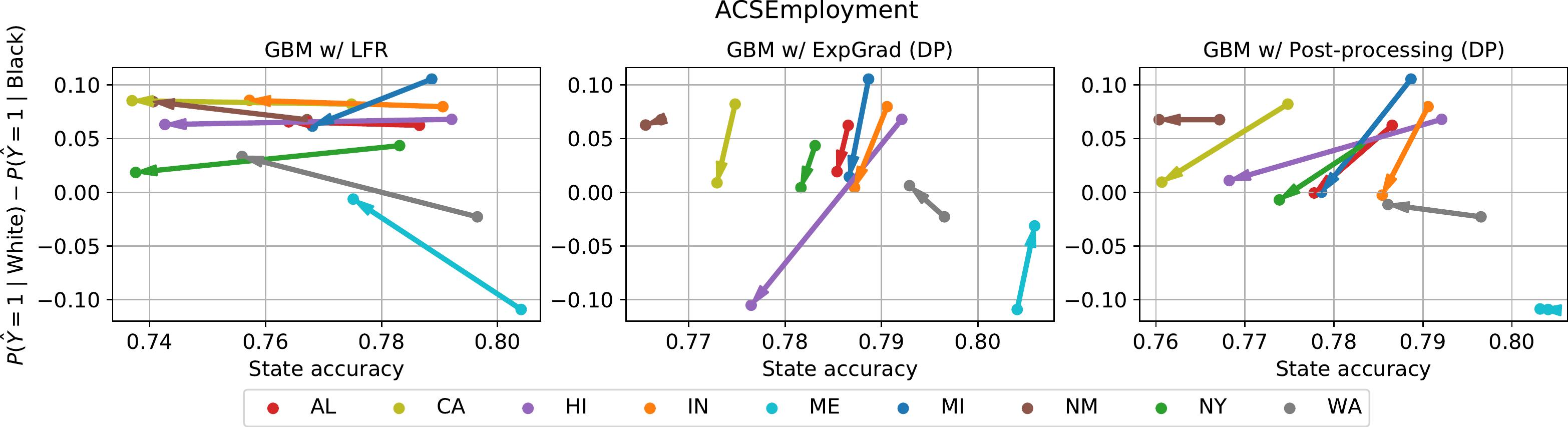}
    \vspace{0.2cm}
    \includegraphics[width=\linewidth]{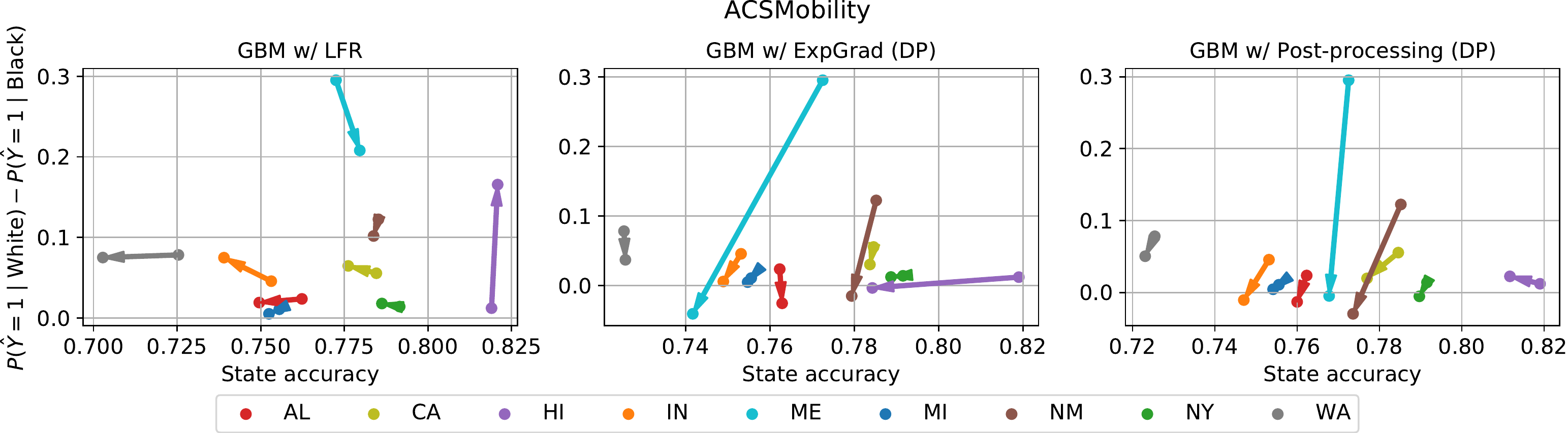}
    \vspace{0.2cm}
    \includegraphics[width=\linewidth]{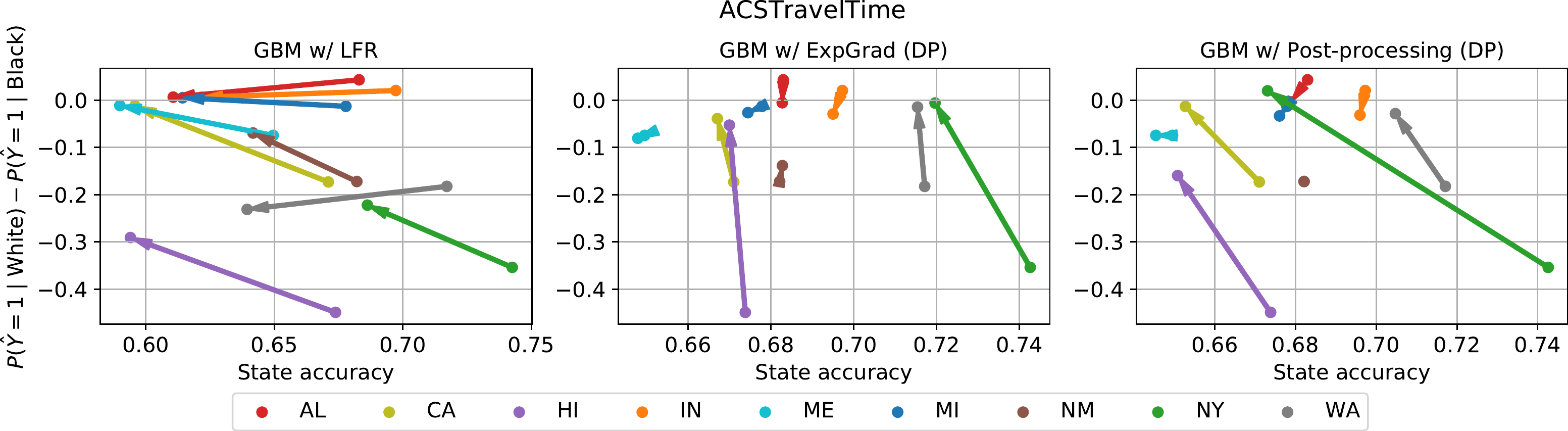}
    \caption{
        The effect size of fairness interventions varies by state. Each panel
        shows the change in accuracy and demographic parity violation (DP) on
        the ACSIncome task after applying a fairness intervention to an
        unconstrained gradient boosted decision tree (GBM). Each arrow
        corresponds to a different state distribution. The arrow base represents
        the (accuracy, DP) point corresponding to the unconstrained GBM, and the
        head represents the (accuracy, DP) point obtained after applying the
        intervention. When only a single point is visible, the entire arrow is
        covered by the point, representing an intervention that has essentially
        no effect.
    }
    \label{fig:fairness_flow_acs}
\end{figure*}

In Figure~\ref{fig:fairness_flow_acs}, we conduct the same experiment for
demographic parity on four other ACS data tasks: ACSPublicCoverage,
ACSEmployment, ACSMobility, and ACSTravelTime, respectively.

\subsection{Geographic distribution shift}
\begin{figure*}
    \centering
    \includegraphics[width=\linewidth]{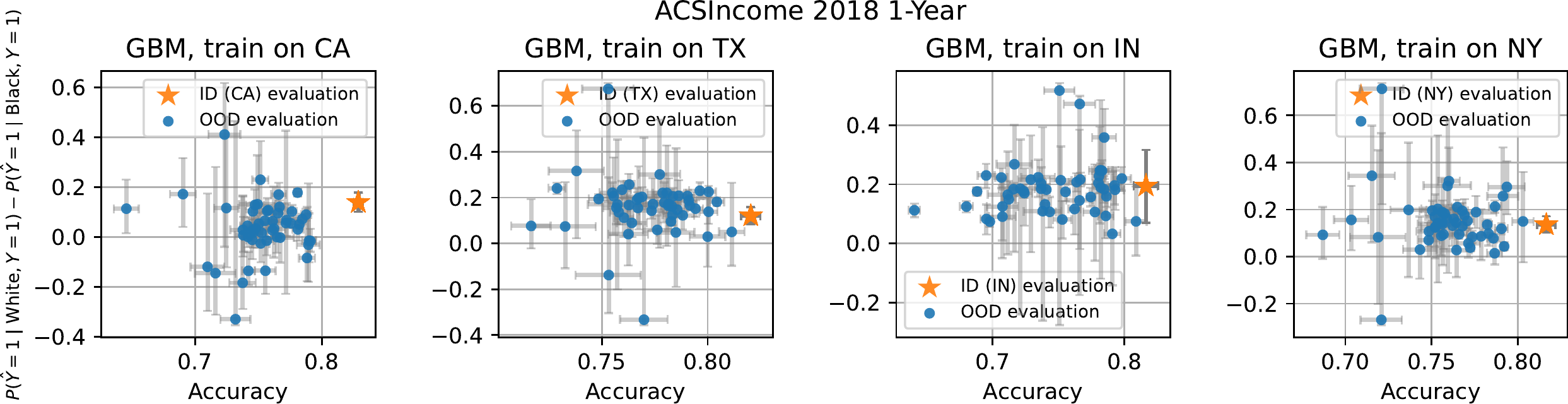}
    \includegraphics[width=\linewidth]{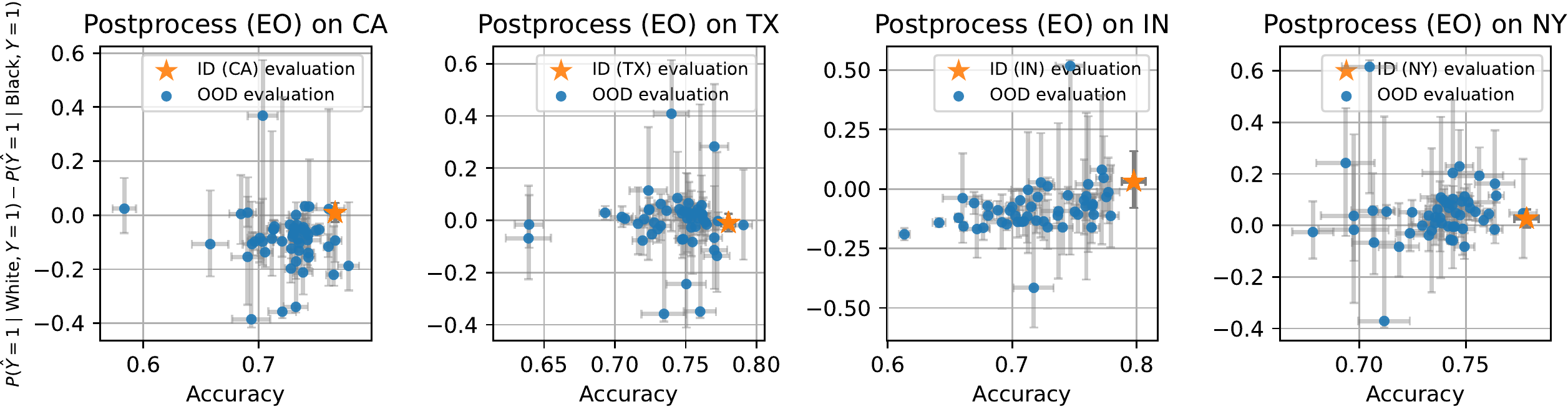}
    \caption{
        Transfer from one state to another gives unpredictable results in terms
        of predictive accuracy and fairness criteria. \textbf{Top:} Each panel
        shows an unconstrained GBM trained on a particular state on the
        ACSIncome task and evaluated both in-distribution (ID) on the same
        state and out-of-distribution (OOD) on the 49 other states in terms of
        accuracy and equality of opportunity violation. \textbf{Bottom:} Each panel
        shows an GBM with post-processing to enforce equality of opportunity
        on the state on which it was trained and evaluated both ID and OOD on
        all 50 states. Confidence intervals are 95\% Clopper-Pearson intervals for accuracy 
        and 95\% Newcombe intervals for equality of opportunity violation.
    }
    \label{fig:pumsadult_state_transfer_eo}
\end{figure*}

\begin{figure*}
    \centering
    \includegraphics[width=\linewidth]{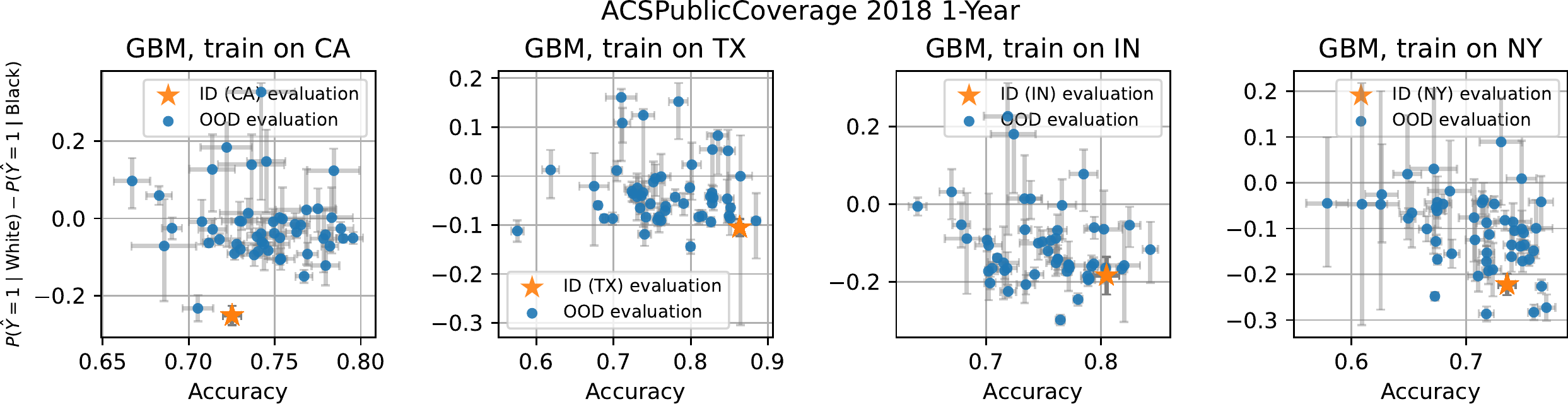}
    \includegraphics[width=\linewidth]{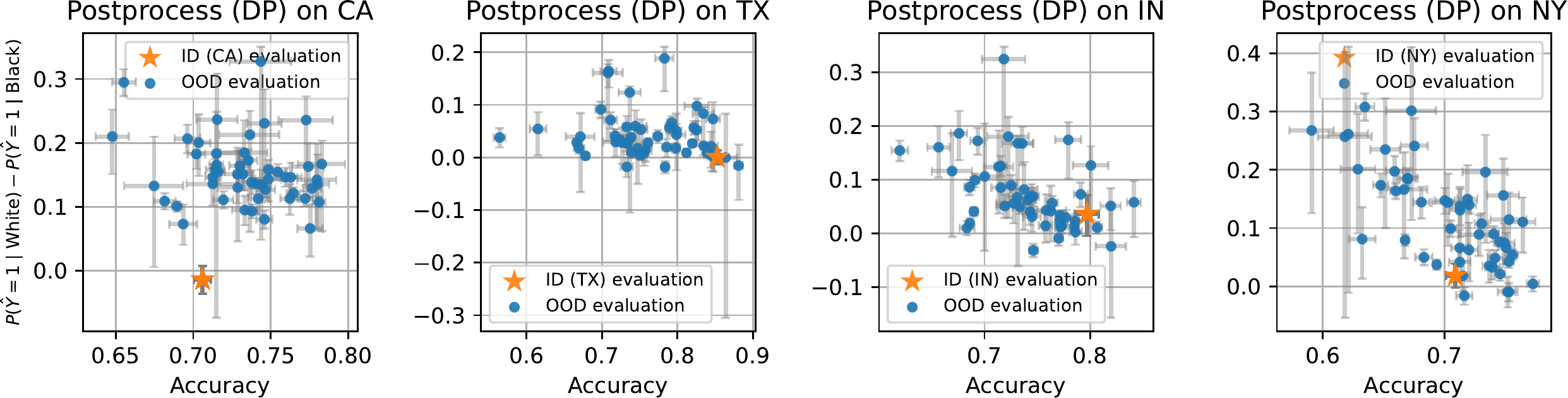}
    \caption{
        Transfer from one state to another gives unpredictable results in terms
        of predictive accuracy and fairness criteria. \textbf{Top:} Each panel
        shows an unconstrained GBM trained on a particular state on the
        ACSPublicCoverage task and evaluated both in-distribution (ID) on the same
        state and out-of-distribution (OOD) on the 49 other states in terms of
        accuracy and equality of opportunity violation. \textbf{Bottom:} Each panel
        shows an GBM with post-processing to enforce equality of opportunity
        on the state on which it was trained and evaluated both ID and OOD on
        all 50 states. Confidence intervals are 95\% Clopper-Pearson intervals for accuracy 
        and 95\% Newcombe intervals for demographic parity.
    }
    \label{fig:publiccoverage_state_transfer}
\end{figure*}

\begin{figure*}
    \centering
    \includegraphics[width=\linewidth]{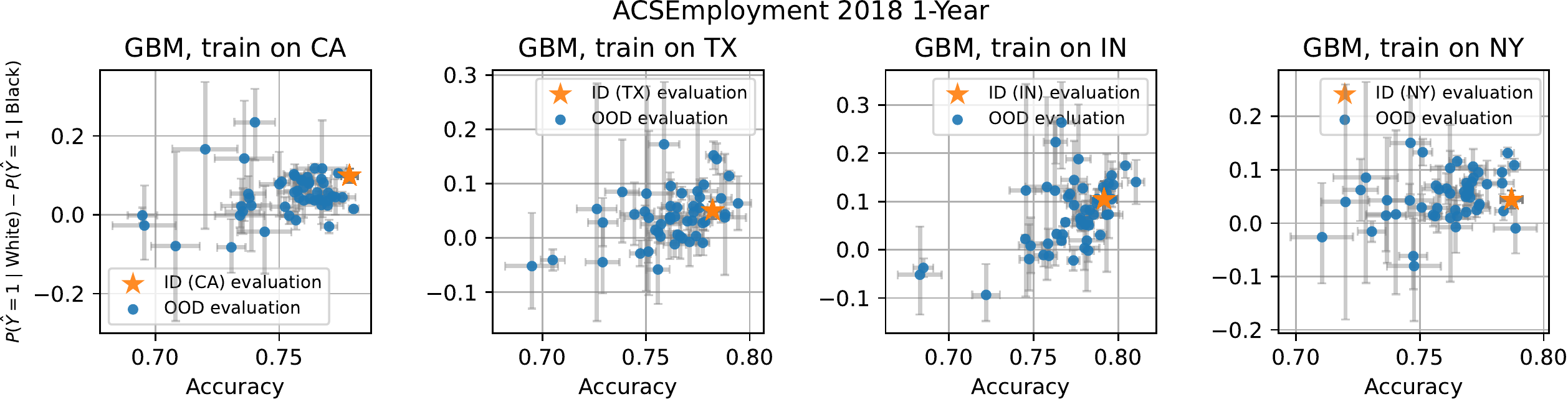}
    \includegraphics[width=\linewidth]{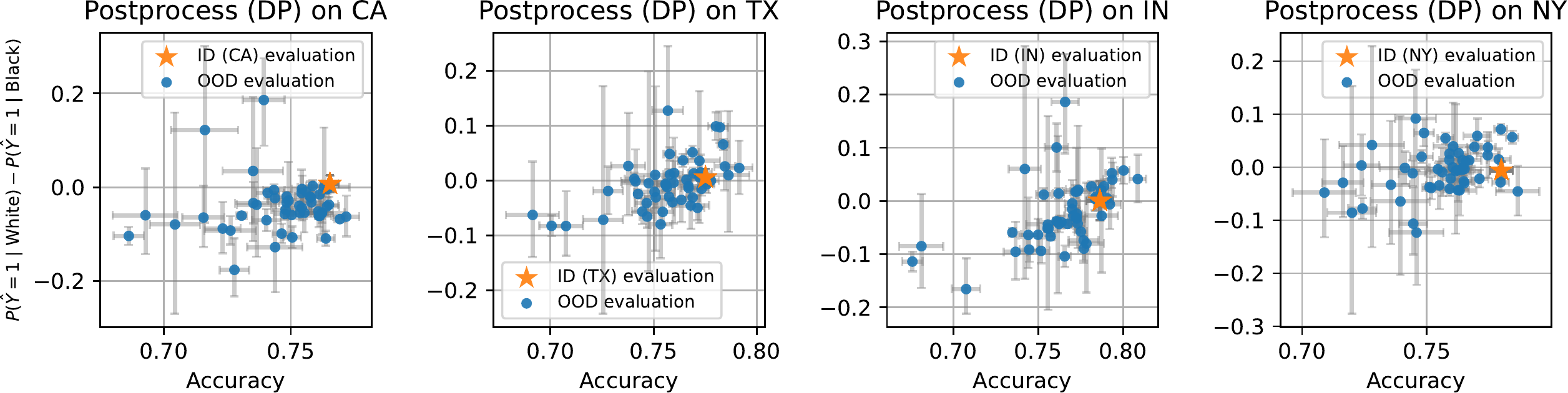}
    \caption{
        Transfer from one state to another gives unpredictable results in terms
        of predictive accuracy and fairness criteria. \textbf{Top:} Each panel
        shows an unconstrained GBM trained on a particular state on the
        ACSEmployment task and evaluated both in-distribution (ID) on the same
        state and out-of-distribution (OOD) on the 49 other states in terms of
        accuracy and equality of opportunity violation. \textbf{Bottom:} Each panel
        shows an GBM with post-processing to enforce equality of opportunity
        on the state on which it was trained and evaluated both ID and OOD on
        all 50 states. Confidence intervals are 95\% Clopper-Pearson intervals for accuracy 
        and 95\% Newcombe intervals for demographic parity.
    }
    \label{fig:employment_state_transfer}
\end{figure*}

\begin{figure*}
    \centering
    \includegraphics[width=\linewidth]{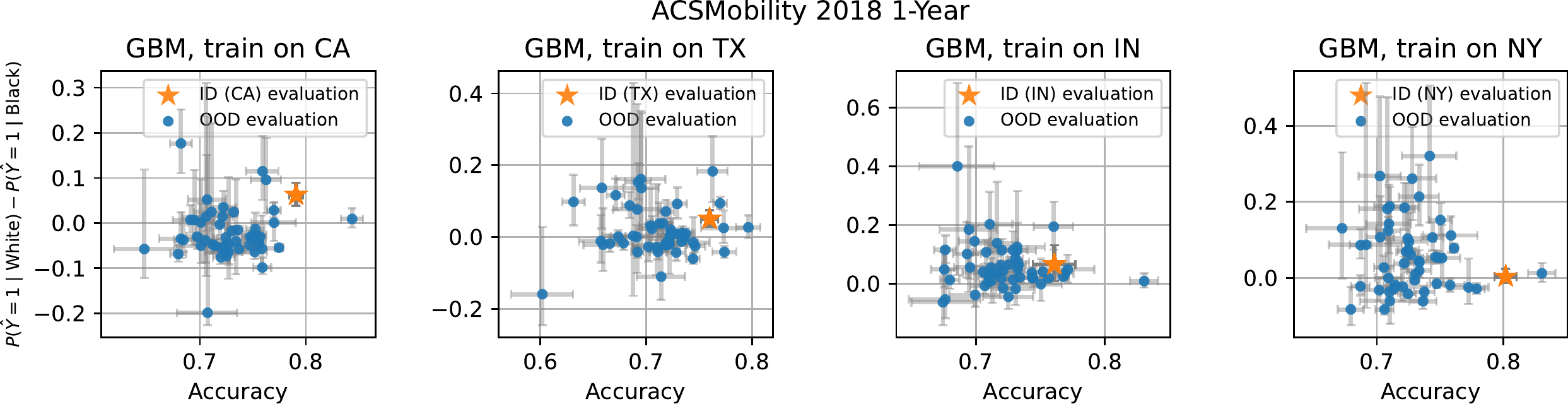}
    \includegraphics[width=\linewidth]{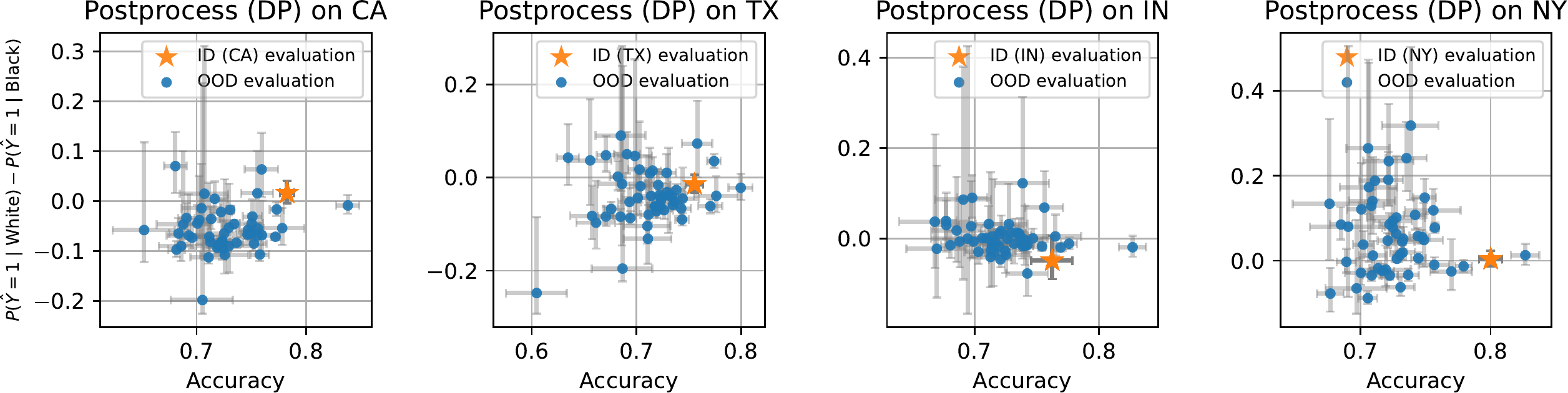}
    \caption{
        Transfer from one state to another gives unpredictable results in terms
        of predictive accuracy and fairness criteria. \textbf{Top:} Each panel
        shows an unconstrained GBM trained on a particular state on the
        ACSMobility task and evaluated both in-distribution (ID) on the same
        state and out-of-distribution (OOD) on the 49 other states in terms of
        accuracy and equality of opportunity violation. \textbf{Bottom:} Each panel
        shows an GBM with post-processing to enforce equality of opportunity
        on the state on which it was trained and evaluated both ID and OOD on
        all 50 states. Confidence intervals are 95\% Clopper-Pearson intervals for accuracy 
        and 95\% Newcombe intervals for demographic parity.
    }
    \label{fig:mobility_state_transfer}
\end{figure*}

\begin{figure*}
    \centering
    \includegraphics[width=\linewidth]{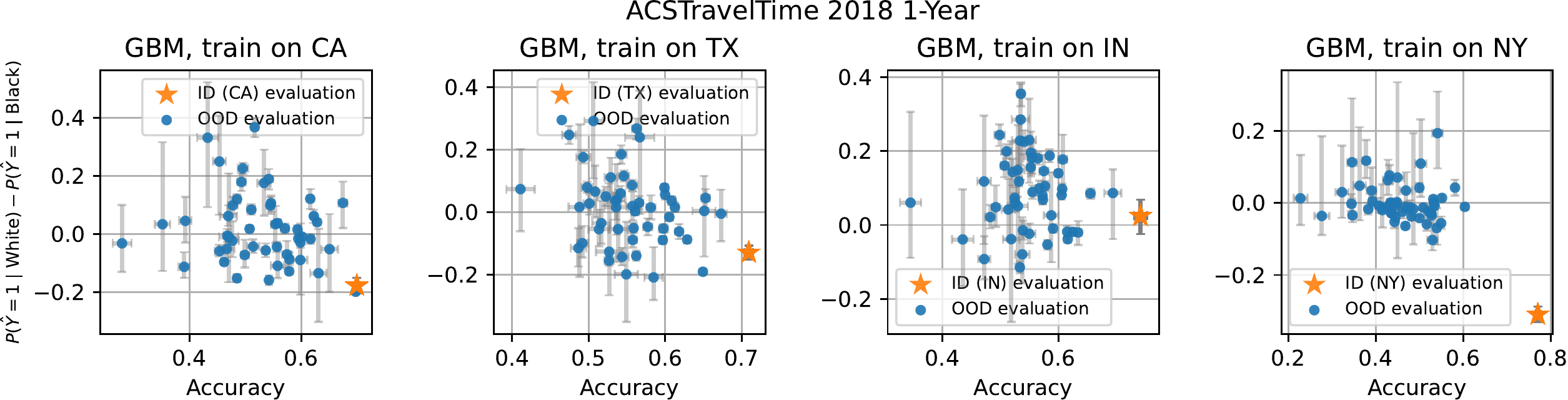}
    \includegraphics[width=\linewidth]{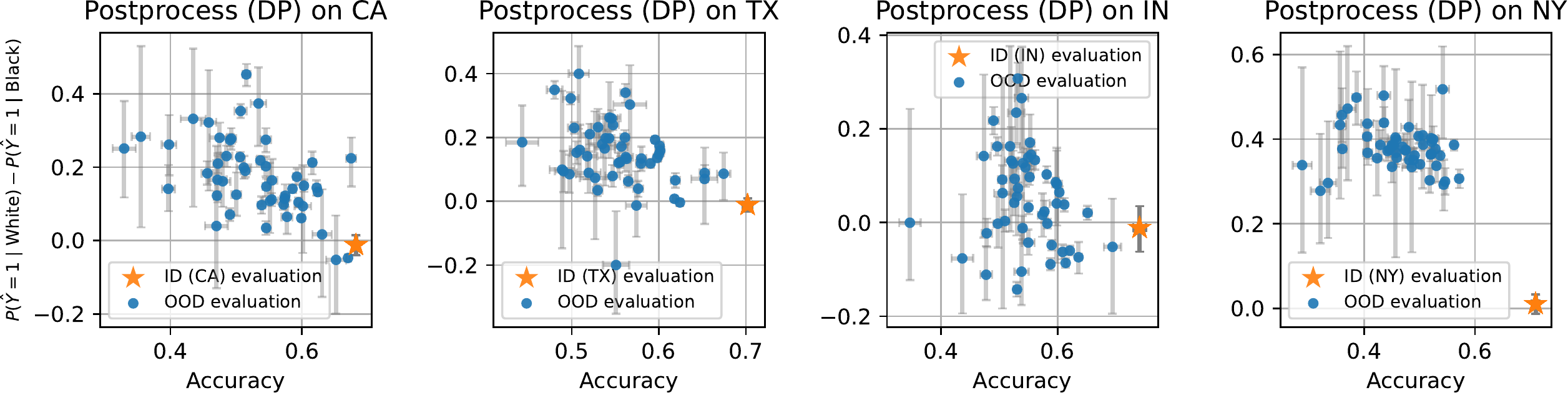}
    \caption{
        Transfer from one state to another gives unpredictable results in terms
        of predictive accuracy and fairness criteria. \textbf{Top:} Each panel
        shows an unconstrained GBM trained on a particular state on the
        ACSTravelTime task and evaluated both in-distribution (ID) on the same
        state and out-of-distribution (OOD) on the 49 other states in terms of
        accuracy and equality of opportunity violation. \textbf{Bottom:} Each panel
        shows an GBM with post-processing to enforce equality of opportunity
        on the state on which it was trained and evaluated both ID and OOD on
        all 50 states. Confidence intervals are 95\% Clopper-Pearson intervals for accuracy 
        and 95\% Newcombe intervals for demographic parity.
    }
    \label{fig:traveltime_state_transfer}
\end{figure*}

In Figure~\ref{fig:pumsadult_state_transfer_eo}, we plot accuracy and
equality of opportunity violation with respect to race for both an unconstrained
GBM and the same model after applying a post-processing adjustment to achieve
equality of opportunity on a natural suite of test sets: the in-distribution (same state test set) and the
out-of-distribution test sets for the 49 other states. This is the same
experiment as in Section~\ref{sec:experiments}, but with equality of opportunity
rather than demographic parity as the metric of interest.
In Figures~\ref{fig:publiccoverage_state_transfer},~\ref{fig:employment_state_transfer}
~\ref{fig:mobility_state_transfer}, and~\ref{fig:traveltime_state_transfer} we
conduct the same experiment for demographic parity on four other ACS data tasks:
ACSPublicCoverage, ACSEmployment, ACSMobility, and ACSTravelTime, respectively.

\subsection{Temporal distribution shift}
\begin{figure*}
    \centering
    \includegraphics[width=\linewidth]{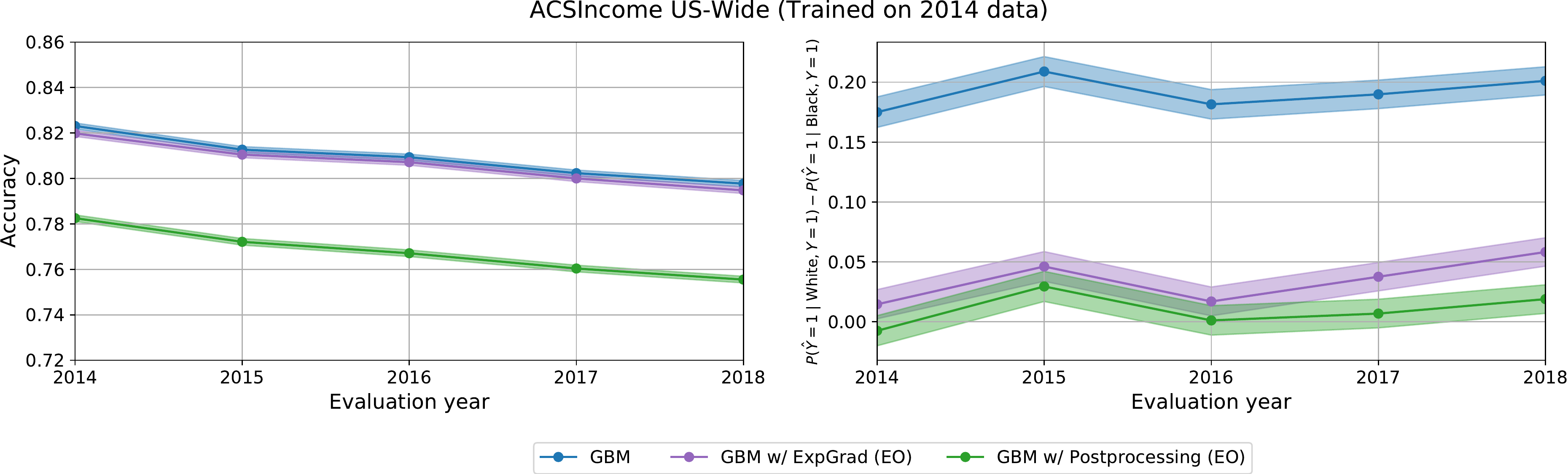}
    \caption{
        Fairness criteria are more stable over time than accuracy.
        \textbf{Left:} Models trained in 2014 on US-wide ACSIncome with and
        without fairness interventions to achieve equality of opportunity and
        evaluated on data in subsequent years.
        \textbf{Right:} Violations of equality of opportunity for the same
        collection of models.  Although accuracy drops over time for most
        problems, violations of equality of opportunity remain essentially
        constant.  Confidence intervals are 95\% Clopper-Pearson intervals for
        accuracy and 95\% Newcombe intervals for equality of opportunity
        violations.
    }
    \label{fig:timedecay_eo}
\end{figure*}

\begin{figure*}
    \centering
    \includegraphics[width=\linewidth]{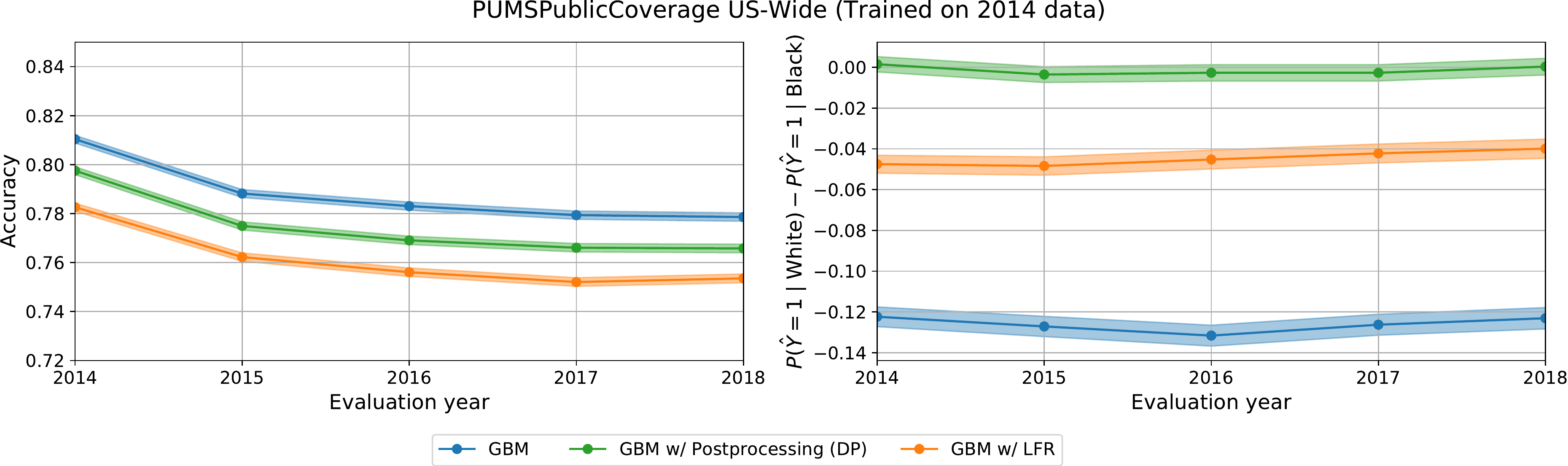}
    \vspace{0.2cm}
    \includegraphics[width=\linewidth]{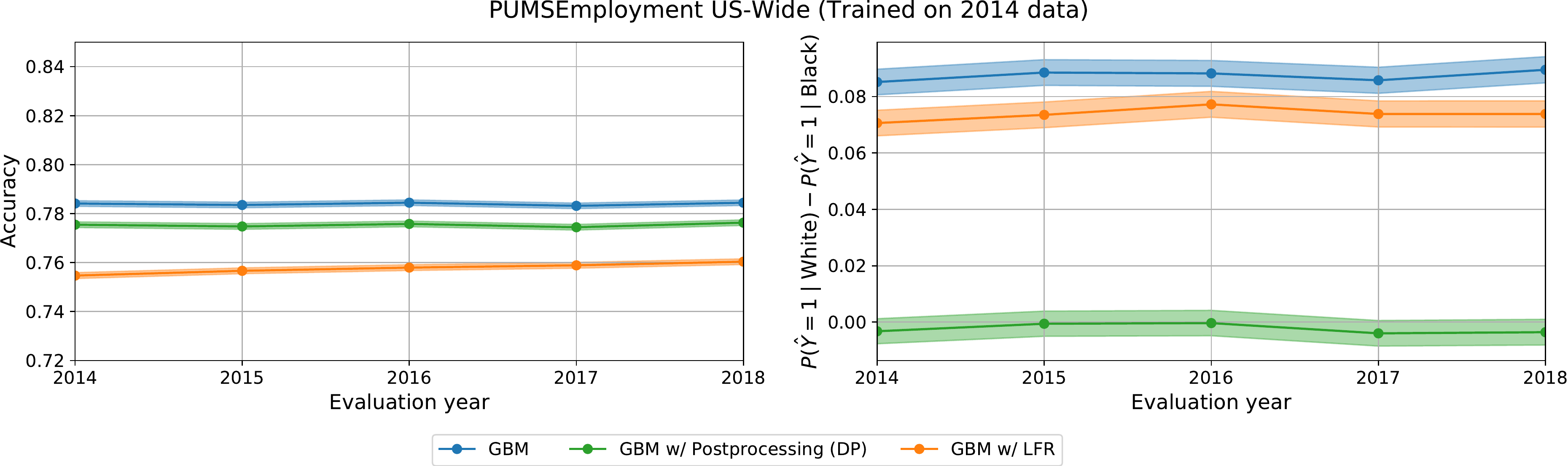}
    \vspace{0.2cm}
    \includegraphics[width=\linewidth]{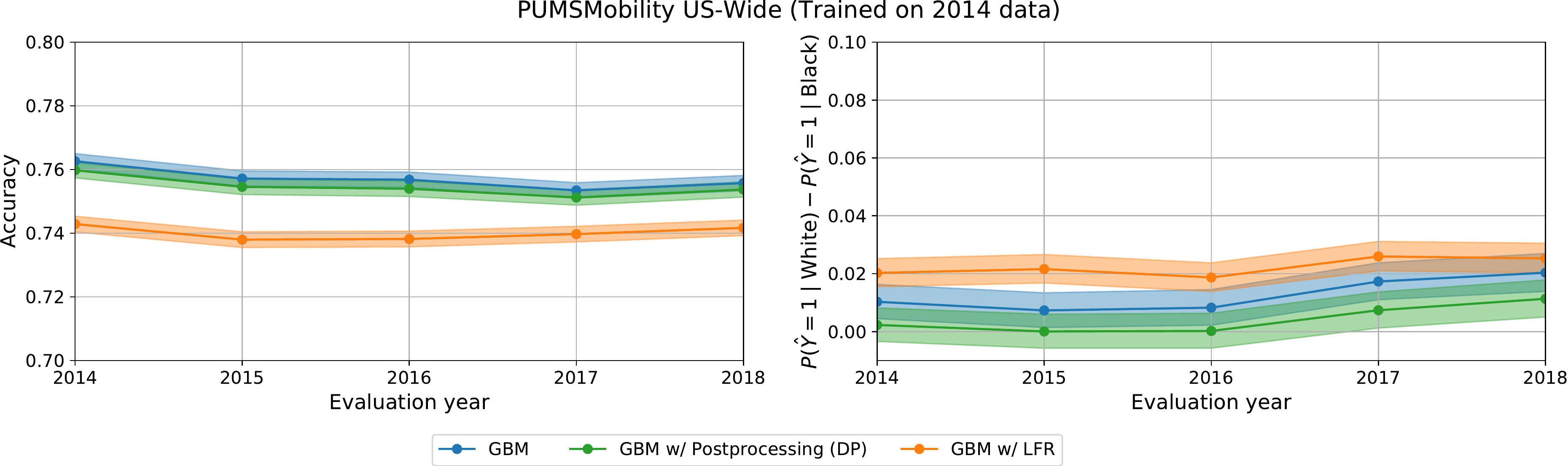}
    \vspace{0.2cm}
    \includegraphics[width=\linewidth]{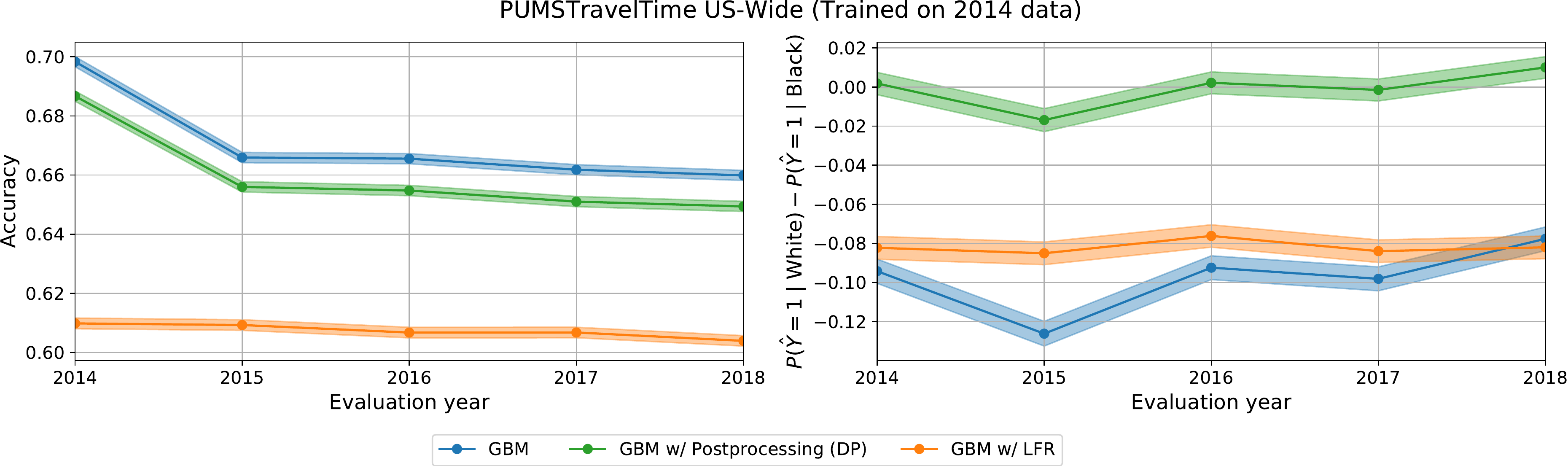}
    \caption{
        Fairness criteria are more stable over time than accuracy.
        \textbf{Left:} Models trained in 2014 on US-wide ACS data with and
        without fairness interventions to achieve demographic parity and
        evaluated on data in subsequent years.
        \textbf{Right:} Violations of demographic parity for the same collection
        of models.
        Although accuracy drops over time for most problems, violations of
        demographic parity remain essentially constant.  Confidence intervals
        are 95\% Clopper-Pearson intervals for accuracy and 95\% Newcombe
        intervals for demographic parity.
    }
    \label{fig:timedecay_dp}
\end{figure*}

In Figure~\ref{fig:timedecay_eo}, we plot model accuracy and
equality of opportunity violation for a GBM trained on the ACSIncome task using
US-wide data from 2014 and evaluated on the test sets for the same task drawn
from years 2014-2018. This is the same experiment as conducted in
Section~\ref{sec:experiments}; however, here we consider interventions to
satisfy equality of opportunity rather than demographic parity.
In Figure~\ref{fig:timedecay_dp}, we conduct repeat this experiment for
interventions to satisfy demographic parity on 4 other ACS PUMS predictions
tasks: ACSPublicCoverage, ACSMobility, ACSEmployment, and ACSTravelTime.

\FloatBarrier

\section{Datasheet}
\label{sec:Datasheet}

This datasheet covers both the prediction tasks we introduce and the underlying US Census data sources. However, due to the extensive documentation available about the US Census data we often point to relevant available resources rather than recreating them here.
For the most up-to-date version of this datasheet, please refer to~\url{https://github.com/zykls/folktables/blob/main/datasheet.md}.

\subsection{Motivation}
\begin{itemize}[leftmargin=*]
\item \textbf{For what purpose was the dataset created?} Was there a specific task
in mind? Was there a specific gap that needed to be filled? Please provide
a description.

The motivation for creating prediction tasks on top of US Census data was to extend the dataset ecosystem available for algorithmic fairness research as outlined in this paper.

\item \textbf{Who created the dataset (e.g., which team, research group) and on
behalf of which entity (e.g., company, institution, organization)?}

The new prediction tasks were created from available US Census data sources by Frances Ding, Moritz Hardt, John Miller, and Ludwig Schmidt.

\item \textbf{Who funded the creation of the dataset?} If there is an associated
grant, please provide the name of the grantor and the grant name and
number.

Frances Ding, Moritz Hardt, and John Miller were employed by the University of California for the duration of this research project, funded by grants administered through the University of California. Ludwig Schmidt was employed by Toyota Research throughout this research project.

\item \textbf{Any other comments?}

No.

\end{itemize}

\subsection{Composition}

\begin{itemize}[leftmargin=*]
    \item \textbf{What do the instances that comprise the dataset represent (e.g.,
    documents, photos, people, countries)?} Are there multiple types of
    instances (e.g., movies, users, and ratings; people and interactions between them; nodes and edges)? Please provide a description.

    Each instance in our IPUMS Adult reconstruction represents an individual.
    Similarly, our datasets derived from ACS contains instances representing individuals.
    The ACS data our datasets are derived from also contain household-level information and the relationship between households and individuals.

    \item \textbf{How many instances are there in total (of each type, if appropriate)?}
   
    Our IPUMS Adult reconstruction contains 49,531 rows (see Section \ref{sec:reconstruction}).
    Table \ref{table:tasks} contains the sizes of our datasets derived from ACS.

    \item \textbf{Does the dataset contain all possible instances or is it a sample
    (not necessarily random) of instances from a larger set?} If the
    dataset is a sample, then what is the larger set? Is the sample representative of the larger set (e.g., geographic coverage)? If so, please describe how
    this representativeness was validated/verified. If it is not representative
    of the larger set, please describe why not (e.g., to cover a more diverse
    range of instances, because instances were withheld or unavailable)

    Both IPUMS Adult and our ACS datasets are samples of the US population.
    Please see Sections \ref{sec:reconstruction} \& \ref{sec:new_data} and the corresponding documentation provided by the US Census Bureau.
    Note that the per-instance weights have to be taken into account if the sample is meant to represent the US population.

    \item \textbf{What data does each instance consist of?} ``Raw'' data (e.g., unprocessed text or images) or features? In either case, please provide a description.

    Each instance consists of features.
    IPUMS Adult uses the same features as the original UCI Adult dataset.
    Appendix \ref{appendix:new-tasks} describes each feature in our new datasets derived from ACS.

    \item \textbf{Is there a label or target associated with each instance?} If so, please
    provide a description.

    Similar to UCI Adult, our IPUMS Adult reconstruction uses the income as label (where the continuous values as opposed to only the binarized values are now available).
    Appendix \ref{appendix:new-tasks} describes the labels in our new datasets derived from ACS.

    \item \textbf{Is any information missing from individual instances?} If so, please
    provide a description, explaining why this information is missing (e.g.,
    because it was unavailable). This does not include intentionally removed
    information, but might include, e.g., redacted text.

    Some features (e.g., the country of origin in IPUMS Adult) contain missing values.
    We again refer to the respective documentation from the US Census Bureau for details.

    \item \textbf{Are relationships between individual instances made explicit
    (e.g., users’ movie ratings, social network links)?}  If so, please describe how these relationships are made explicit.

    Our versions of the datasets contain no relationships between individuals.
    The original data sources from the US Census contain relationships between individuals and households.

    \item \textbf{Are there recommended data splits (e.g., training, development/validation,
    testing)?}  If so, please provide a description of these splits, explaining
    the rationale behind them.

    For IPUMS Adult, it is possible to follow the same train / test split as the original UCI Adult.
    In general, we recommend k-fold cross-validation for all of our datasets.

    \item \textbf{Are there any errors, sources of noise, or redundancies in the
dataset?} If so, please provide a description.

    Our IPUMS Adult reconstruction contains slightly more rows than the original UCI Adult, see Section \ref{sec:reconstruction}.
    Beyond IPUMS Adult, we refer to the documentation of CPS and ACS provided by the US Census Bureau.

    \item \textbf{Is the dataset self-contained, or does it link to or otherwise rely on
    external resources (e.g., websites, tweets, other datasets)? } If it links
    to or relies on external resources, a) are there guarantees that they will
    exist, and remain constant, over time; b) are there official archival versions
    of the complete dataset (i.e., including the external resources as they
    existed at the time the dataset was created); c) are there any restrictions
    (e.g., licenses, fees) associated with any of the external resources that
    might apply to a future user? Please provide descriptions of all external
    resources and any restrictions associated with them, as well as links or
    other access points, as appropriate.

    Due to restrictions on the re-distribution of the original IPUMS and ACS data sources, we do not provide our datasets as standalone data files.
    Instead, we provide scripts to generate our datasets from the respective sources.

    Both the US Census Bureau and IPUMS aim to provide stable long-term access to their data.
    Hence we consider these data sources to be reliable.
    We refer to the IPUMS website and the website of the US Census Bureau for specific usage restrictions.
    Neither data source has fees associated with it.

    \item \textbf{Does the dataset contain data that might be considered confidential (e.g., data that is protected by legal privilege or by doctor patient confidentiality, data that includes the content of individuals’ non-public communications)?} If so, please provide a description.

    Our datasets are subsets of datasets released publicly by the US Census Bureau.

    \item \textbf{ Does the dataset contain data that, if viewed directly, might be offensive, insulting, threatening, or might otherwise cause anxiety?} If so, please describe why.

    No.

    \item \textbf{Does the dataset relate to people?} If not, you may skip the remaining
    questions in this section.

    Yes, each instance in our datasets corresponds to a person.

    \item \textbf{Does the dataset identify any subpopulations (e.g., by age, gender)?} If so, please describe how these subpopulations are identified and
    provide a description of their respective distributions within the dataset.

    Our datasets identify subpopulations since each individual has features such as age, gender, or race.
    Please see the main text of our paper for experiments exploring the respective distributions.

    \item \textbf{Is it possible to identify individuals (i.e., one or more natural persons), either directly or indirectly (i.e., in combination with other
    data) from the dataset?}  If so, please describe how.

    To the best of our knowledge, it is not possible to identify individuals \emph{directly} from our datasets.
    However, the possibility of reconstruction attacks combining data from the US Cenus Bureau (such as CPS and ACS) and other data sources are a concern and actively investigated by the research community.

    \item \textbf{Does the dataset contain data that might be considered sensitive
    in any way (e.g., data that reveals racial or ethnic origins, sexual
    orientations, religious beliefs, political opinions or union memberships, or locations; financial or health data; biometric or genetic data; forms of government identification, such as social security numbers; criminal history)?}  If so, please provide a description.

    Our datasets contain features such as race, age, or gender that are often considered sensitive.
    This is by design since we assembled our datasets to test algorithmic fairness interventions.

    \item \textbf{Any other comments?}
    
    No.

\end{itemize}

\subsection{Collection process}

    \begin{itemize}[leftmargin=*]
    \item \textbf{How was the data associated with each instance acquired?} Was
the data directly observable (e.g., raw text, movie ratings), reported by
subjects (e.g., survey responses), or indirectly inferred/derived from other
data (e.g., part-of-speech tags, model-based guesses for age or language)?
If data was reported by subjects or indirectly inferred/derived from other
data, was the data validated/verified? If so, please describe how.
    
    The data was reported by subjects as part of the ACS and CPS surveys.
    The respective documentation provided by the US Census Bureau contains further information, see \url{https://www.census.gov/programs-surveys/acs/methodology/design-and-methodology.html} and \url{https://www.census.gov/programs-surveys/cps/technical-documentation/methodology.html}.

    \item \textbf{What mechanisms or procedures were used to collect the data (e.g., hardware apparatus or sensor, manual human curation, software program, software API)?}
    How were these mechanisms or procedures validated?

    The ACS relies on a combination of internet, mail, telephone, and in-person interviews.
    CPS uses in-person and telephone interviews.
    Please see the aforementioned documentation from the US Census Bureau for detailed information.

    \item \textbf{If the dataset is a sample from a larger set, what was the sampling strategy (e.g., deterministic, probabilistic with specific sampling probabilities)?}
    
    For the ACS, the US Census Bureau sampled housing units uniformly for each county.
    See Chapter 4 in the ACS documentation (\url{https://www2.census.gov/programs-surveys/acs/methodology/design_and_methodology/acs_design_methodology_report_2014.pdf}) for details.

    CPS is also sampled by housing unit from certain sampling areas, see Chapters 3 and 4 in \url{https://www.census.gov/prod/2006pubs/tp-66.pdf}.

    \item \textbf{Who was involved in the data collection process (e.g., students, crowdworkers, contractors) and how were they compensated (e.g., how much were crowdworkers paid)?}

    The US Census Bureau employs interviewers for conducting surveys.
    According to online job information platforms such as \url{indeed.com}, an interviewer earns about \$15 per hour.

    \item \textbf{Over what timeframe was the data collected?} Does this timeframe match the creation timeframe of the data associated with the instances (e.g., recent crawl of old news articles)? If not, please describe the timeframe in which the data associated with the instances was created.
    
    Both CPS and ACS collect data annually.
    Our IPUMS Adult reconstruction contains data from the 1994 CPS ASEC.
    Our new tasks derived from ACS can be instantiated for various survey years.

    \item \textbf{Were any ethical review processes conducted (e.g., by an institutional review board)?}
    If so, please provide a description of these review processes, including the outcomes, as well as a link or other access point to any supporting documentation.
    
    Both ACS and CPS are regularly reviewed by the US Census Bureau.
    As a government agency, the US Census Bureau is also subject to government oversight mechanisms.

    \item \textbf{Does the dataset relate to people? If not, you may skip the remainder of the questions in this section.}

Yes.

    \item \textbf{Did you collect the data from the individuals in question directly, or obtain it via third parties or other sources (e.g., websites)?}

Data collection was performed by the US Census Bureau. We obtained the data from publicly available US Census repositories.

    \item \textbf{Were the individuals in question notified about the data collection?} If so, please describe (or show with screenshots or other information) how notice was provided, and provide a link or other access point
to, or otherwise reproduce, the exact language of the notification itself.

Yes. A sample ACS form is available online: \url{https://www.census.gov/programs-surveys/acs/about/forms-and-instructions/2021-form.html}

Information about the CPS collection methodology is available here: \url{https://www.census.gov/programs-surveys/cps/technical-documentation/methodology.html}

    \item \textbf{Did the individuals in question consent to the collection and use of their data?}
    If so, please describe (or show with screenshots or other information) how consent was requested and provided, and provide a link or other access point to, or otherwise reproduce, the exact language to which the individuals consented.

Participation in the US Census American Community Survey is mandatory. Participation in the US Corrent Population Survey is voluntary and consent is obtained at the beginning of the interview: \url{https://www2.census.gov/programs-surveys/cps/methodology/CPS-Tech-Paper-77.pdf}

    \item \textbf{If consent was obtained, were the consenting individuals provided with a mechanism to revoke their consent in the future or for certain uses?}
    If so, please provide a description, as well as a link or other access point to the mechanism (if appropriate).

We are not aware that the Census Bureau would provide such a mechanism.

    \item \textbf{Has an analysis of the potential impact of the dataset and its use on data subjects (e.g., a data protection impact analysis) been conducted?}
    If so, please provide a description of this analysis, including the outcomes, as well as a link or other access point to any supporting documentation.

The US Census Bureau assesses privacy risks and invests in statistical disclosure control. See \url{https://www.census.gov/topics/research/disclosure-avoidance.html}. Our derived prediction tasks do not increase privacy risks.

    \item \textbf{Any other comments?}

    No.
    \end{itemize}

\subsection{Preprocessing / cleaning / labeling}
\begin{itemize}[leftmargin=*]
	\item \textbf{Was any preprocessing/cleaning/labeling of the data done (e.g.,
discretization or bucketing, tokenization, part-of-speech tagging,
SIFT feature extraction, removal of instances, processing of missing values)?} If so, please provide a description. If not, you may skip the
remainder of the questions in this section.

We used two US Census data products -- we reconstructed UCI Adult from the Annual Social and Economic Supplement (ASEC) of the Current Population Survey (CPS), and we constructed new prediction tasks from the American Community Survey (ACS) Public Use Microdata Sample (PUMS). Before releasing CPS data publicly,  the Census Bureau top-codes certain variables and conducts imputation of certain missing values, as documented here: \url{https://www.census.gov/programs-surveys/cps/technical-documentation/methodology.html}. In our IPUMS Adult reconstruction, we include a subset of the variables available from the CPS data and do not alter their values.

The ACS data release similarly top-codes certain variables and conducts imputation of certain missing values, as documented here: \url{https://www.census.gov/programs-surveys/acs/microdata/documentation.html}. For the new prediction tasks that we define, we further process the ACS data as documented at the folktables GitHub page,
\url{https://github.com/zykls/folktables}.
In most cases, this involves mapping missing values (NaNs) to $-1$. We release code so that new prediction tasks may be defined on the ACS data, with potentially different preprocessing. Each prediction task also defines a binary label by discretizing the target variable into two classes; this can be easily changed to define a new labeling in a new prediction task.

\item \textbf{Was the ``raw'' data saved in addition to the preprocessed/cleaned/labeled
data (e.g., to support unanticipated future uses)?} If so, please provide a link or other access point to the ``raw'' data.

Yes, our package provides access to the data as released by the U.S. Census Bureau. The ``raw'' survey answers collected by the Census Bureau are not available for public release due to privacy considerations.

\item \textbf{Is the software used to preprocess/clean/label the instances available?} If so, please provide a link or other access point.

The software to is available at the folktables GitHub page,
\url{https://github.com/zykls/folktables}.

\item \textbf{Any other comments?}

No.

\end{itemize}

\subsection{Uses}
\begin{itemize}[leftmargin=*]
	\item \textbf{Has the dataset been used for any tasks already?} If so, please provide a description.
	
	In this paper we create five new prediction tasks from the ACS PUMS data: 
	\begin{enumerate}
		\item ACSIncome: Predict whether US working adults' yearly income is above \$50,000.
		\item ACSPublicCoverage: Predict whether a low-income individual, not eligible for Medicare, has coverage from public health insurance.
		\item ACSMobility: Predict whether a young adult moved addresses in the last year.
		\item ACSEmployment: Predict whether a US adult is employed.
		\item ACSTravelTime: Predict whether a working adult has a travel time to work of greater than 20 minutes. 

	\end{enumerate}

Further details about these tasks can be found at the folktables GitHub page,
\url{https://github.com/zykls/folktables},
and in Appendix \ref{appendix:new-tasks}.

	\item \textbf{Is there a repository that links to any or all papers or systems that use the dataset?} If so, please provide a link or other access point.
	
	At the folktables GitHub page,
\url{https://github.com/zykls/folktables}, any public forks to the package are visible, and papers or systems that use the datasets should cite the paper linked at that Github page.
	
	\item \textbf{What (other) tasks could the dataset be used for?}
	
	New prediction tasks may be defined on the ACS PUMS data that use different subsets of variables as features and/or different target variables. Different prediction tasks may have different properties such as Bayes error rate, or the base rate disparities between subgroups, that can help to benchmark machine learning models in diverse settings.  
	
	\item \textbf{Is there anything about the composition of the dataset or the way
it was collected and preprocessed/cleaned/labeled that might impact future uses?}
For example, is there anything that a future user
might need to know to avoid uses that could result in unfair treatment
of individuals or groups (e.g., stereotyping, quality of service issues) or
other undesirable harms (e.g., financial harms, legal risks) If so, please
provide a description. Is there anything a future user could do to mitigate
these undesirable harms?

Both the CPS and ACS are collected through surveys of a subset of the US population, and in their documentation, they acknowledge that statistical trends in individual states may be noisy compared to those found by analyzing US data as a whole, due to small sample sizes in certain states. In particular, there may be very few individuals with particular characteristics (e.g. ethnicity) in certain states, and generalizing conclusions from these few individuals may be highly inaccurate. Further, benchmarking fair machine learning algorithms on datasets with few representatives of certain subgroups may provide the illusion of ``checking a box'' for fairness, without substantive merit.

\item \textbf{Are there tasks for which the dataset should not be used?}
If so,
please provide a description.

This dataset contains personal information, and users should not attempt to re-identify individuals in it. Further, these datasets are meant primarily to aid in benchmarking machine learning algorithms; Census data is often crucial for substantive, domain-specific work by social scientists, but our dataset contributions are not in this direction. Substantive investigations into inequality, demographic shifts, and other important questions should not be based purely on the datasets we provide. 

\item \textbf{Any other comments?}

No.

\end{itemize}

\subsection{Distribution}
\begin{itemize}[leftmargin=*]
\item \textbf{Will the dataset be distributed to third parties outside of the entity (e.g., company, institution, organization) on behalf of which the dataset was created?}
If so, please provide a description.

The dataset will be available for public download on the folktables GitHub page,
\url{https://github.com/zykls/folktables}.

\item \textbf{How will the dataset will be distributed (e.g., tarball on website, API, GitHub)?}
Does the dataset have a digital object identifier (DOI)?

The dataset will be be distributed via GitHub, see \url{https://github.com/zykls/folktables}. The dataset does not have a DOI.

\item \textbf{When will the dataset be distributed?}

The dataset will be released on August 1, 2021 and available thereafter for
download and public use.

\item \textbf{Will the dataset be distributed under a copyright or other intellectual property (IP) license, and/or under applicable terms of use (ToU)?}
If so, please describe this license and/or ToU, and provide a link
or other access point to, or otherwise reproduce, any relevant licensing
terms or ToU, as well as any fees associated with these restrictions.

The folktables package and data loading code will be available under the MIT
license. The folktables data itself is based on data from the American Community
Survey (ACS) Public Use Microdata Sample (PUMS) files managed by the US Census
Bureau, and it is governed by the terms of use provided by the Census Bureau.
For more information, see
\url{https://www.census.gov/data/developers/about/terms-of-service.html}

Similarly, the IPUMS adult reconstruction is governed by the IPUMS terms of use.
For more information, see \url{https://ipums.org/about/terms}.

\item \textbf{Have any third parties imposed IP-based or other restrictions on the data associated with the instances?}
If so, please describe these restrictions, and provide a link or other access point to, or otherwise reproduce, any relevant licensing terms, as well as any fees associated with these restrictions.

The folktables data and the adult reconstruction data are governed by
third-party terms of use provided by the US Census Bureau and IPUMS,
respectively. See
\url{https://www.census.gov/data/developers/about/terms-of-service.html} and
\url{https://ipums.org/about/terms} for complete details.
The IPUMS Adult Reconstruction is a subsample of the IPUMS CPS data available
from \url{cps.ipums.org} These data are intended for replication purposes only.
Individuals analyzing the data for other purposes must submit a separate data
extract request directly via IPUMS CPS. Individuals should contact
\url{ipums@umn.edu} for redistribution requests.

\item \textbf{Do any export controls or other regulatory restrictions apply to the dataset or to individual instances?}
If so, please describe these restrictions, and provide a link or other access point to, or otherwise reproduce, any supporting documentation.

To our knowledge, no export controls or regulatory restrictions apply to the
dataset.

\item \textbf{Any other comments?}

No.
\end{itemize}

\subsection{Maintenance}

\begin{itemize}[leftmargin=*]
\item \textbf{Who is supporting/hosting/maintaining the dataset?}

The dataset will be hosted on GitHub, and supported and maintained by
the folktables team. As of June 2021, this team consists of Frances Ding, Moritz
Hardt, John Miller, and Ludwig Schmidt.

\item \textbf{How can the owner/curator/manager of the dataset be contacted
(e.g., email address)?}

Please send issues and requests to \url{folktables@gmail.com}.

\item \textbf{Is there an erratum?} If so, please provide a link or other access point.

An erratum will be hosted on the dataset website,
\url{https://github.com/zykls/folktables}.

\item \textbf{Will the dataset be updated (e.g., to correct labeling errors, add
new instances, delete instances)?} If so, please describe how often, by
whom, and how updates will be communicated to users (e.g., mailing list,
GitHub)?

The dataset will be updated as required to address errors and refine the
prediction problems based on feedback from the community. The package
maintainers will update the dataset and communicate these updates on GitHub.

\item \textbf{If the dataset relates to people, are there applicable limits on the
retention of the data associated with the instances (e.g., were individuals in question told that their data would be retained for a fixed period of time and then deleted)?}
If so, please describe these limits and explain how they will be enforced.

The data used in folktables is based on data from the American Community Survey
(ACS) Public Use Microdata Sample (PUMS) files managed by the US Census Bureau.
The data inherits and will respect the corresponding retention policies of the
ACS. Please see \url{https://www.census.gov/programs-surveys/acs/about.html} for
more details.  For the Adult reconstruction dataset, the data is based on
Current Population Survey (CPS) released by IPUMS and thus inherits and will
respect the corresponding retention policies for the CPS. Please see
\url{https://cps.ipums.org/cps/} for more details. 

\item \textbf{Will older versions of the dataset continue to be
supported/hosted/maintained?} If so, please describe how. If not, please describe how its obsolescence
will be communicated to users.

Older versions of the datasets in folktables will be clearly indicated,
supported, and maintained on the GitHub website. Each new version of the dataset
will be tagged with {\tt version} metadata and an associated GitHub release.

\item \textbf{If others want to extend/augment/build on/contribute to the
dataset, is there a mechanism for them to do so?} If so, please
provide a description. Will these contributions be validated/verified?
If so, please describe how. If not, why not? Is there a process for communicating/distributing these contributions to other users? If so, please
provide a description.

Users wishing to contribute to folktables datasets are encouraged to do so by
submitting a pull request on the website
\url{https://github.com/zykls/folktables/pulls}. The contributions will be
reviewed by the maintainers. These contributions will be reflected in new
version of the dataset and broadcasted as part of each Github release.

\item \textbf{Any other comments?}

No.
\end{itemize}

\end{document}